\documentclass[10pt,twocolumn,letterpaper]{article}

\usepackage{cvpr}              

\usepackage{graphicx}
\usepackage{amsmath}
\usepackage{amssymb}
\usepackage{booktabs}
\usepackage{bbding}

\usepackage[pagebackref,breaklinks,colorlinks]{hyperref}

\usepackage[capitalize]{cleveref}
\crefname{section}{Sec.}{Secs.}
\Crefname{section}{Section}{Sections}
\Crefname{table}{Table}{Tables}
\crefname{table}{Tab.}{Tabs.}

\def\cvprPaperID{*****} 
\def\confName{CVPR}
\def\confYear{2022}

\begin{document}

\title{Deep Models with Fusion Strategies \\ for MVP  Point Cloud Registration} 

\author{Lifa Zhu\\
DeepGlint\\
Beijing, China\\
{\tt\small lifazhu@deepglint.com}
\and
Changwei Lin\\
DeepGlint\\
Beijing, China\\
{\tt\small changweilin@deepglint.com}
\and
Dongrui Liu\\
Shanghai Jiao Tong University\\
Shanghai, China\\
{\tt\small drliu96@sjtu.edu.cn}
\and
Xin Li\\
Sichuan University \\
Chengdu, China\\
{\tt\small lixinscu@163.com}
\and
Francisco Gómez-Fernández\\
DeepGlint\\
Beijing, China\\
{\tt\small francisco@deepglint.com}
}

\maketitle

\section{Introduction}
\label{sec:intro}

The main goal of point cloud registration in Multi-View Partial (MVP) Challenge 2021\footnote{\url{https://competitions.codalab.org/competitions/33430}}~\cite{pan2021variational} is to estimate a rigid transformation to align a point cloud pair. The pairs in this competition have the characteristics of low overlap, non-uniform density, unrestricted rotations and ambiguity, which pose a huge challenge to the registration task.

In this report, we introduce our solution to the registration task, which fuses two deep learning models: ROPNet and PREDATOR~\cite{huang2020registration}, with customized ensemble strategies.
We propose ROPNet, a new deep learning model using Representative Overlapping Points with discriminative features for registration that transforms partial-to-partial registration into partial-to-complete registration. 
Specifically, we propose a context-guided module which uses an encoder to extract global features for predicting point overlap score. 
To better find representative overlapping points, we use the extracted global features for coarse alignment. 
Then, we introduce a Transformer~\cite{guo2020pct} to enrich point features and remove non-representative points based on point overlap score and feature matching. 
A similarity matrix is built in a partial-to-complete mode, and finally, weighted SVD is adopted to estimate a transformation matrix.
%

PREDATOR ~\cite{huang2020registration} is a recent work which implements pairwise point cloud registration with deep attention on overlap region. It shows great performance in both 3DMatch~\cite{zeng20173dmatch} and 3DLowMatch~\cite{huang2020registration}. However, object-centric point clouds in MVP registration are different from scene-centric  data in 3DMatch, which are ambiguous or symmetric. (detailed information can be seen in \autoref{sec:ensemble}). That is why we chose an ensemble model  strategy. Furthermore, to make PREDATOR work better, we solved a simple but important GNN bug\footnote{\url{https://github.com/overlappredator/OverlapPredator/issues/15}}, adjusted parameters for the MVP registration challenge and also, solved registration in a partial-to-complete manner during RANSAC iterations.

Finally, we propose a few ensemble rules based on data characteristics to fuse ROPNet and PREDATOR that help to achieve a better performance in a large variety of cases.
We achieved 3.16656, 0.029237 and 0.08451 on the validation set with the metrics of Rot\_Error, Trans\_Error and MSE, respectively. 
In the MVP Point Cloud Challenge 2021 we achieved the $2^\text{nd}$ place in the registration track with error values of $2.96546$, $0.02632$ and $0.07808$ on test set.

\section{Team Details}

\begin{itemize}

\item[$\bullet$] CodaLab user name

zhulf0804

\item[$\bullet$] Team name

miscellaneous

\item[$\bullet$] Team member names

Lifa Zhu, Changwei Lin, Dongrui Liu, Xin Li, Francisco Gómez-Fernández

\item[$\bullet$] Affiliation(s)

Deep Glint, Shanghai Jiaotong University, Sichuan University

\item[$\bullet$] CodaLab email address

lifazhu@deepglint.com



\item[$\bullet$] Final rank of the team in the development phase

$2^\text{nd}$ in the registration track.

\item[$\bullet$] Link to the codes/executables of the solution(s) 

We released our ROPNet implementation at \url{https://github.com/zhulf0804/ROPNet}. The official code of PREDATOR was released by the the authors at \url{https://github.com/overlappredator/OverlapPredator}. We also released our unofficial implementation of PREDATOR at \url{https://github.com/zhulf0804/PREDATOR}.

\end{itemize}

\section{Contribution Details}
\begin{itemize}

\item[$\bullet$] Title of the contribution

Deep models with fusion strategies for partial-to-partial point cloud registration under unrestricted rotations.

\item[$\bullet$] General method description

We propose to fuse ROPNet and PREDATOR\cite{huang2020registration} to solve point cloud registration in the MVP Challenge. ROPNet and the pipeline of our solution to this challenge can be seen in \autoref{fig:ROPNet} and  \autoref{fig:pipeline}.

\begin{figure*}
\centering
\includegraphics[width=0.99\linewidth]{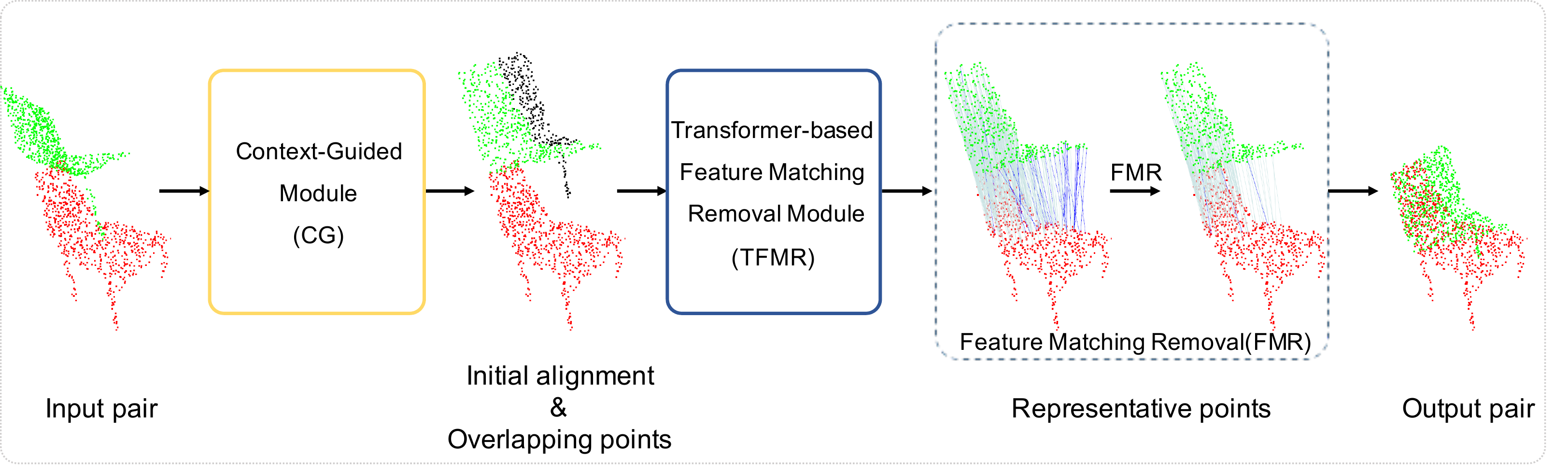}
\caption{Overview of ROPNet registration pipeline. The CG module consumes the source (green) and target (red) point clouds, and outputs initial pose and overlapping points (non-overlapping points are in black). The TFMR module takes the output of CG module as input, and generates accurate correspondences.
The FMR step removes false correspondences (blue lines) and keeps some positive correspondences (gray lines).}
\label{fig:ROPNet}
\end{figure*}

\begin{figure*}[t]
\begin{center}
   \includegraphics[width=0.8\linewidth]{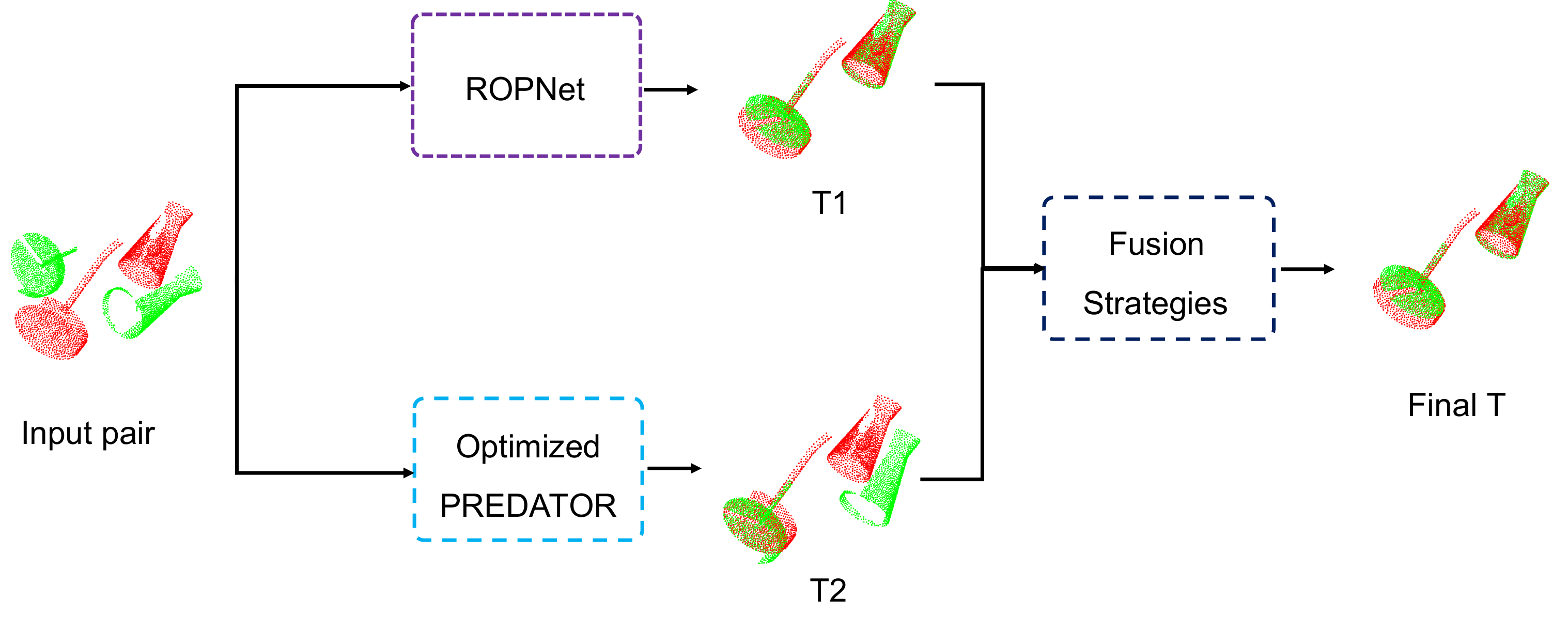}
\end{center}
   \caption{The pipeline of our solution to the MVP Registration Challenge. The point cloud in green is the source point cloud,  and the point cloud in red is the target point cloud.}
\label{fig:pipeline}
\end{figure*}

In ROPNet, we proposed a context-guided module which uses an encoder to extract global features for predicting point overlap score and introduced a Transformer to enrich point features and remove non-representative points based on point overlap score and feature matching. Using ROPNet, we achieved low rotation and translation error on validation set whose rot\_level=0, as shown in \autoref{table:single}.

\begin{figure*}[t]
\centering
\includegraphics[width=0.9\textwidth]{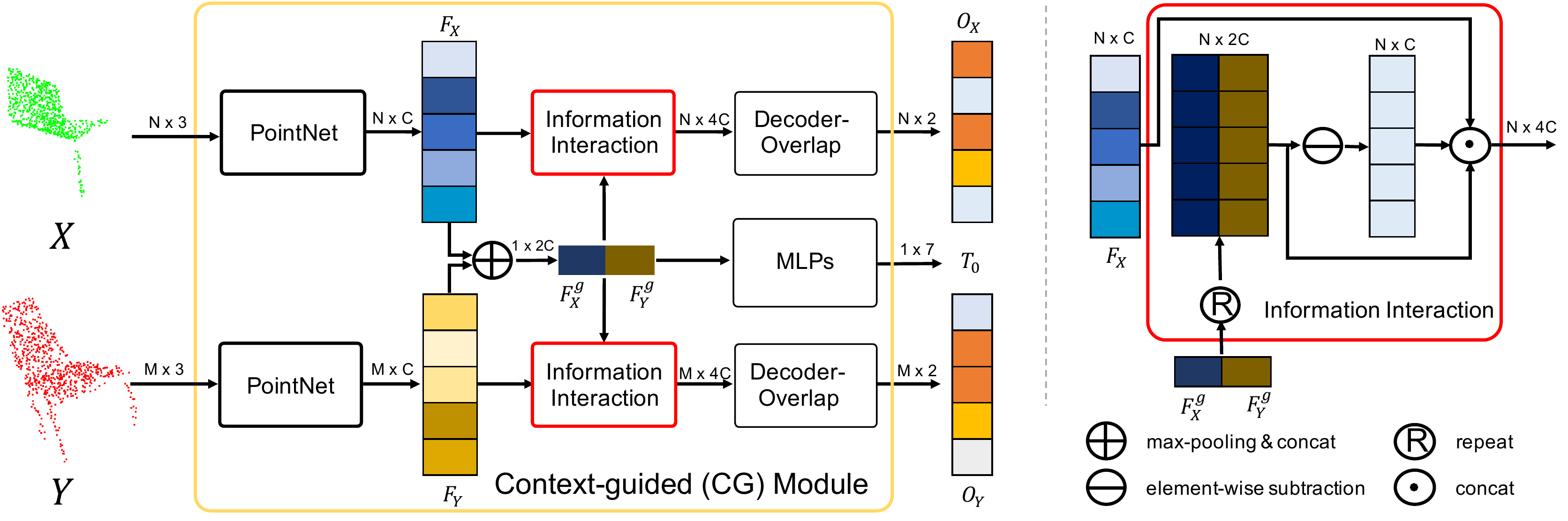} \\
\caption{Left: Architecture of the CG module. CG module consumes source $X$ (in green) and target $Y$ (in red) data, and outputs overlap score ($O_X$, $O_Y$) and initial transformation matrix $T_{0}$. Right: Details of information interaction. It takes point features and global features as input and outputs fused point features based on the pair.}
\label{fig:ROP_CG}
\end{figure*}

\begin{figure*}[t]
\centering
\includegraphics[width=0.9\textwidth]{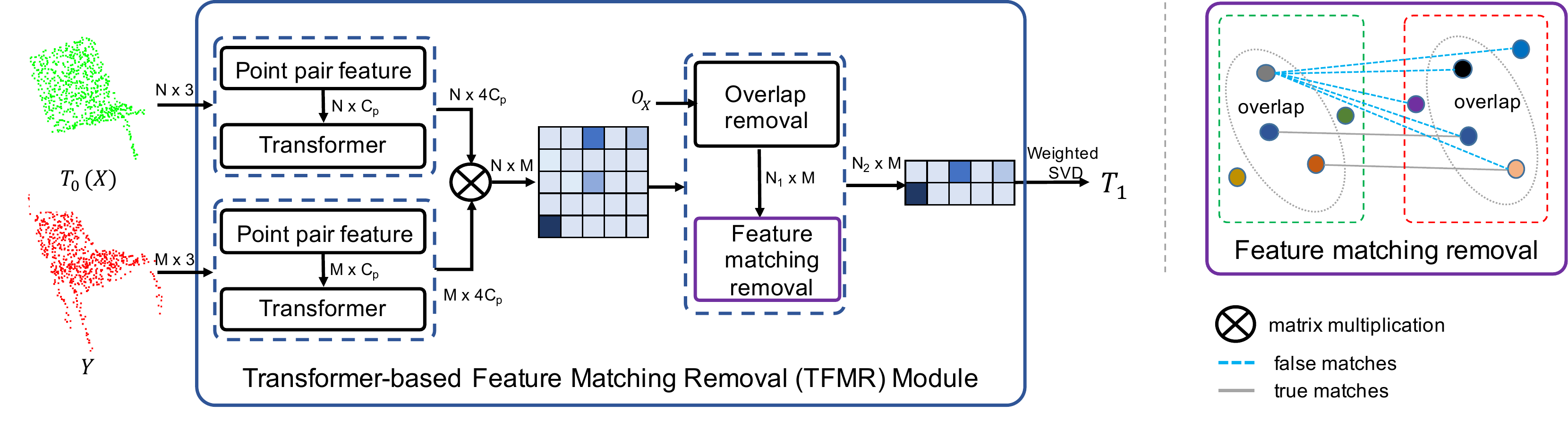} \\
\caption{Left: Overview of Transformer-based feature matching removal (TFMR) module. TFMR module takes the transformed source $X'$ and target $Y$ as input, and outputs representative points and their correspondences. $T_0$ and $O_X$ are the output from CG module that denote initial alignment and overlap score for $X$.
Right: Details of feature matching removal (FMR). It takes correspondences for overlapping source points and outputs accurate correspondences (gray lines).}
\label{fig:ROP_TFMR}
\end{figure*}

Considering unrestricted rotations, we used PREDATOR in our pipeline.
In PREDATOR source code, we found and solved a simple but important GNN bug which helps the network to obtain a higher performance.
Then, we adjusted parameters in PREDATOR for the MVP registration challenge. Inspired by the idea of partial-to-complete proposed in ROPNet, we try to remove some points in source data based on the predicted scores and keep all points in target data during RANSAC iterations.
However, as shown in the results reported in \autoref{table:single}, the registration error is still not ideal due to the data characteristics. We will analyse the data characteristics in \autoref{sec:ensemble}. 

Based on the above discussions, we designed a few ensemble strategies based on data characteristics to help fuse ROPNet and PREDATOR to estimate the final rigid transformation. 
Experiments on the validation set showed it is effective on most cases, with few fails. 

\item[$\bullet$] Representative image / diagram of the method(s)

ROPNet can be seen in \autoref{fig:ROPNet}.
Our proposed context-guided (CG) module and Transformer-based Feature Matching Removal (TFMR) Module in ROPNet can seen in \autoref{fig:ROP_CG} and \autoref{fig:ROP_TFMR}.
The pipeline of our solution is shown in \autoref{fig:pipeline}.


\end{itemize}

\section{Method and Data Details}
\label{sec:method}
\begin{itemize}



\item[$\bullet$] Training description

We trained ROPNet and PREDATOR independently. All 2048 points were involved in training for the two models.
For ROPNet, we trained for 600 epochs using Adam optimizer with initial learning rate of 0.0001. The learning rate changes using a cosine annealing schedule. 
We trained ROPNet in a non-iterative manner. 
However, we run 2 iterations for the TFMR module during test.
It is noted that we only trained ROPNet for small rotation angles ranging from 0° to 45°. 
Also, we did not use Point Pair Features \cite{drost2010model}, because we could not get accurate normal vectors in the MVP challenge data. 

For PREDATOR, following the code\footnote{\url{https://github.com/overlappredator/OverlapPredator}} released by the authors, we trained on MVP registration dataset for 200 epochs using SGD with 0.98 momentum. The initial learning rate was 0.01, with an exponential decay factor of 0.95 every epoch.
We trained PREDATOR under unrestricted rotation angles ranging from 0° to 360°. In addition, we adjusted some parameters such as voxel size to 0.04, sampled points in circle loss, and others in loss implementation.

\item[$\bullet$] Testing description

For each source and target point cloud pair, we estimate transformations $T_1 \in \mathbf{SE(3)}$ and $T_3 \in \mathbf{SE(3)}$ based on ROPNet and PREDATOR, respectively. $T_1$ is the output predicted from source to target using end-to-end ROPNet model. $T_3$ is also the transformation from source to target, which is obtained with RANSAC using features and keypoints provided by PREDATOR. We select $T1$ or $T_3$ based on our proposed ensemble rules, which will be introduced in \autoref{sec:ensemble}.

\item[$\bullet$] Results of the comparison to other approaches

We compare our method with  RPMNet\_corr~\cite{zodage2020correspondence}, which is the variant of RPMNet~\cite{yew2020rpm} to help solving registration with unrestricted rotations. The results in \autoref{table:other_method} shows that our ROPNet achieves much lower registration error than RPMNet\_corr when rot\_level is 0. When rot\_level is not restricted, the ensemble model of ROPNet and PREDATOR also achieves much lower registration error than RPMNet\_corr.

\begin{table}
\begin{center}
\scalebox{0.7}{
\begin{tabular}{|c|c|c|c|c|}
\hline
Model & rot\_level & Error(R) & Error(t) & MSE  \\
\hline
RPMNet\_corr & 0 & 12.5560 & 0.1674 & 0.3865 \\
ROPNet & 0 & 1.0449 & 0.0193 & 0.0375 \\
RPMNet\_corr & 0, 1 & 21.9685 & 0.2062 & 0.5896 \\
ROPNet + PREDATOR &  0, 1 & 3.16656 & 0.029237 & 0.08451 \\
\hline
\end{tabular}}
\end{center}
\caption{Comparison to other approaches on val set.}
\label{table:other_method}
\end{table}

\item[$\bullet$] Results on other standard benchmarks

We compared our ROPNet with several classic deep learning registration networks in the ModelNet40~\cite{wu20153d} dataset, including DCP~\cite{wang2019deep}, IDAM~\cite{li2019iterative}, DeepGMR~\cite{yuan2020deepgmr} and RPMNet~\cite{yew2020rpm}. 
We generate partial point cloud pairs following RPMNet, then we use 40 categories in ModelNet40 for training, and test over 40 categories on the test set. 
We evaluate the registration in terms of the isotropic rotation and translation error $$Error(R) = \text{arccos}\frac{tr(\hat R^{-1}R) - 1}{2}$$, $$Error(t) = ||\hat R^{-1} t - \hat R^{-1} \hat t||_1$$ proposed in RPMNet\cite{yew2020rpm}, where $R, t$ and $\hat R, \hat t$ represent the predicted and the ground truth transformation respectively, $tr(\cdot)$ means the trace of matrix. 
Moreover, we evaluate the isotropic rotation and translation error $MAE(R), MAE(t)$ used in DCP~\cite{wang2019deep} by calculating mean absolute error of Euler angle and translation vector. 
Both $Error(R)$ and $MAE(R)$ represent rotation error in degrees.
The results in \autoref{table:1} indicate that ROPNet outperforms other methods, exceeding DCP, IDAM and DeepGMR by a large margin.
This is also the reason why we chose ROPNet as our baseline method.

\begin{table}
\begin{center}
\scalebox{0.8}{
\begin{tabular}{|l|c|c|c|c|}
\hline
Methods & $Error(R)$ & $Error(t)$ & $MAE(R)$ & $MAE(t)$ \\
\hline
DCP-v2 & 11.1723 & 0.1356 & 5.6421 & 0.0657 \\
IDAM-GNN & 14.2891 & 0.1909 & 7.4966 & 0.0877 \\
DeepGMR & 14.3612 & 0.1589 & 7.0914 & 0.0775 \\
RPMNet & 1.4239 & 0.0139 & 0.7304 & 0.0065 \\
ROPNet & {\bf 1.1567} & {\bf 0.0108} & {\bf 0.5946} & {\bf 0.0051} \\
\hline
\end{tabular}}
\end{center}
\caption{Results on other standard benchmarks (ModelNet40 unseen shapes).}
\label{table:1}
\end{table}

\item[$\bullet$] Novelty degree of the solution and whether it has been previously published

\begin{itemize}
\item We proposed an end-to-end network ROPNet which may be the first work to transform partial-to-partial registration to partial-to-complete registration. 
\item A simple yet effective CG module is proposed to obtain overlapping points and an initial alignment.
\item We proposed TFMR module which uses transformer to enrich point feature and removes non-representative points by feature matching.
\item We proposed a few customized ensemble strategies to fuse registration networks for MVP registration challenge.
\end{itemize}

Our preprint version of ROPNet can be accessed at \url{https://arxiv.org/pdf/2107.02583.pdf} which has not yet been published.

\item[$\bullet$] Comment the robustness and generality of the proposed solution(s)?

In order to show the robustness and generality of our work, we conducted further experiments that are  discussed in the following. 

To validate the model generalization ability, we use the first 8 categories for training and the rest 8 categories for testing. As shown in \autoref{table:unseen_cat}, both ROPNet and PREDATOR have good generalization abilities. Also, comparing the second and the third row, we can see that our ROPNet has a better generalization than PREDATOR.


We conducted robustness experiments in the first 8 categories mainly considering two aspects: noise and point cloud density. Firstly, we add noise which is sampled from $N(0, 0.01^2)$ and clipped to $[-0.5, 0.5]$ for each point independently to validate the model robustness. Secondly, we trained on 2048 points and evaluate on 1024 points sampled from the original 2048 points. As shown in the top-3 rows of \autoref{table:robustness}, ROPNet is robust to density variation and noise. From the bottom 2 rows, we can see that the ensemble model is robust to noise.

Furthermore, we conducted generalization and robustness experiments on ModelNet40 partial-to-partial registration. 
We use the first 20 categories for training and the rest 20 categories for testing and validating the model generalization ability. We evaluate the robustness of model in the presence of noise sampled in the same way as before. 
As shown in \autoref{table:unseen_cat_modelnet40}, ROPNet is robust to noise with good generalization abilities.

However, there are still some hard cases or ambiguity cases that we can't register them well. The last row in \autoref{fig:results_vis} shows some bad cases in validation set whose rotation errors (in degree) are bigger than 10°.

\begin{table}
\begin{center}
\scalebox{0.8}{
\begin{tabular}{|c|c|c|c|}
\hline
Model & Error(R) & Error(t) & MSE \\
\hline
ROPNet(8) + PREDATOR(8) & 6.0865 & 0.0408 & 0.1470 \\
ROPNet(16) + PREDATOR(8) & 5.5636 & 0.0303 & 0.1274 \\
ROPNet(8) + PREDATOR(16) & 4.1196 & 0.0290 & 0.1009  \\
ROPNet(16) + PREDATOR(16) & 3.8302 & 0.0251 & 0.0920 \\
\hline
\end{tabular}}
\end{center}
\caption{Results on MVP registration unseen categories. The number in brackets indicates the number of training categories.}
\label{table:unseen_cat}
\end{table}

\begin{table}
\begin{center}
\scalebox{0.6}{
\begin{tabular}{|c|c|c|c|c|c|c|}
\hline
Model & rot\_level & Noise & npoints & Error(R) & Error(t) & MSE \\
\hline
ROPNet & 0 & \XSolid & 2048 & 1.2599 & 0.0199 & 0.0418 \\ 
ROPNet & 0 & \XSolid & 1024 & 1.5246 & 0.0264 & 0.0530 \\
ROPNet & 0 & \checkmark & 1024 & 1.6637 & 0.0291 & 0.0581 \\
ROPNet + PREDATOR & 0, 1 & \XSolid & 2048 & 1.9003 & 0.0260 & 0.0592 \\
ROPNet + PREDATOR & 0, 1 & \checkmark & 2048 & 2.4904 & 0.0249 & 0.0684 \\
\hline
\end{tabular}}
\end{center}
\caption{Results on the first 8 categories of MVP to validate the robustness.}
\label{table:robustness}
\end{table}

\begin{table}
\begin{center}
\scalebox{0.8}{
\begin{tabular}{|c|c|c|c|c|c|}
\hline
Methods & Noise & $Error(R)$ & $Error(t)$ & $MAE(R)$ & $MAE(t)$ \\
\hline
ROPNet & \XSolid & 1.1637 & 0.0116 & 0.6190 & 0.0055 \\
ROPNet & \checkmark & 1.4656 & 0.0145 & 0.7799 & 0.0070 \\
\hline
\end{tabular}}
\end{center}
\caption{Results on ModelNet40 unseen categories.}
\label{table:unseen_cat_modelnet40}
\end{table}

\item[$\bullet$] Comment the efficiency of the proposed solution(s)?

We evaluate the inference speed of ROPNet, PREDATOR and ROPNet + PREDATOR, independently. Both the source and target point cloud have 2048 points. Tests were run on a single GeForce GTX TITAN X with Intel(R) Core(TM) i7-6700K CPU @ 4.00GHz, 32GB RAM, which is different from the training environment. As shown in \autoref{table:speed}, we need 0.152s, 0.457s and 0.768s for one pair registration with ROPNet, PREDATOR and ensemble models, respectively. By updating RANSAC iterations in PREDATOR from 2M to 0.1M, the ensemble models are more faster with a slight performance decrease. 

\begin{table}
\begin{center}
\scalebox{0.7}{
\begin{tabular}{|c|c|c|c|}
\hline
Model & rot\_level & time (s) & Error(R)  \\
\hline
ROPNet & 0 & 0.152 &  1.0449 \\
PREDATOR (RANSAC 0.1M) & 0, 1 & 0.270 & 7.6118 \\
PREDATOR (RANSAC 2M) &  0, 1 & 0.457 & 6.8290 \\
ROPNet + PREDATOR (RANSAC 0.1M) &  0, 1 & 0.580 & 3.86217  \\
ROPNet + PREDATOR (RANSAC 2M) &  0, 1 & 0.768 &  3.16656 \\
\hline
\end{tabular}}
\end{center}
\caption{Inference speed for different models.}
\label{table:speed}
\end{table}

\end{itemize}

\section{Ensembles and Fusion Strategies / Ablation Studies (if any)}
\label{sec:ensemble}
\begin{itemize}

\item[$\bullet$] Describe in detail the use of ensembles and/or fusion strategies (if any). 

\begin{figure*}
\scalebox{1.0}{
\begin{tabular}{cccc}
\includegraphics[width=0.24\textwidth]{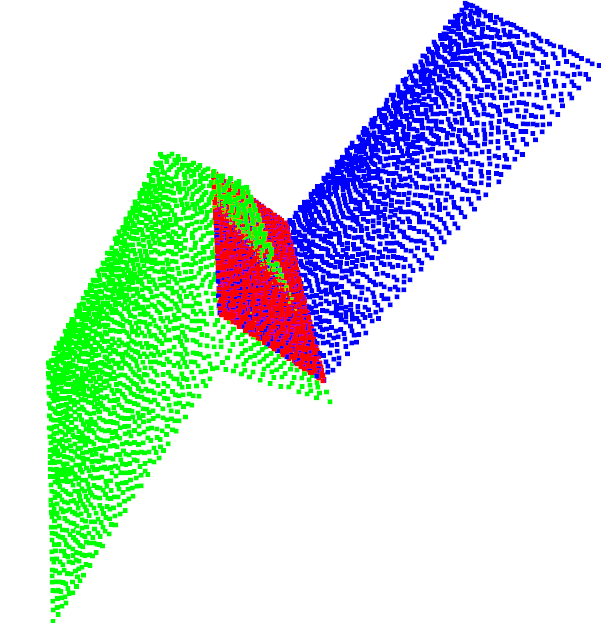} & \includegraphics[width=0.24\textwidth]{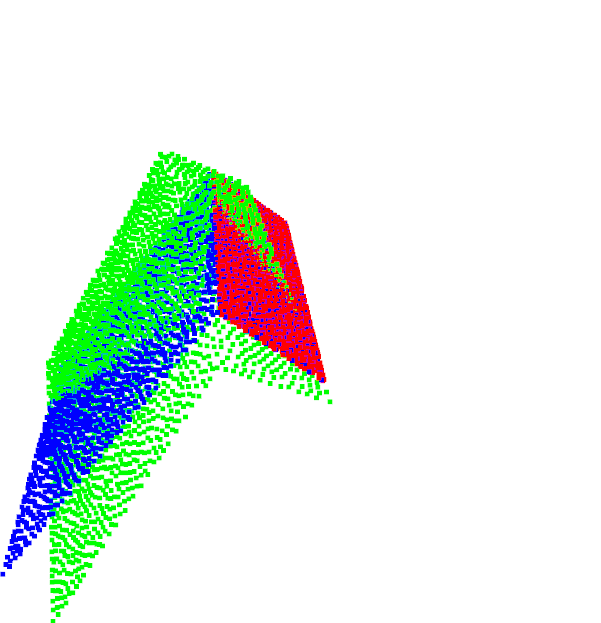} & \includegraphics[width=0.24\textwidth]{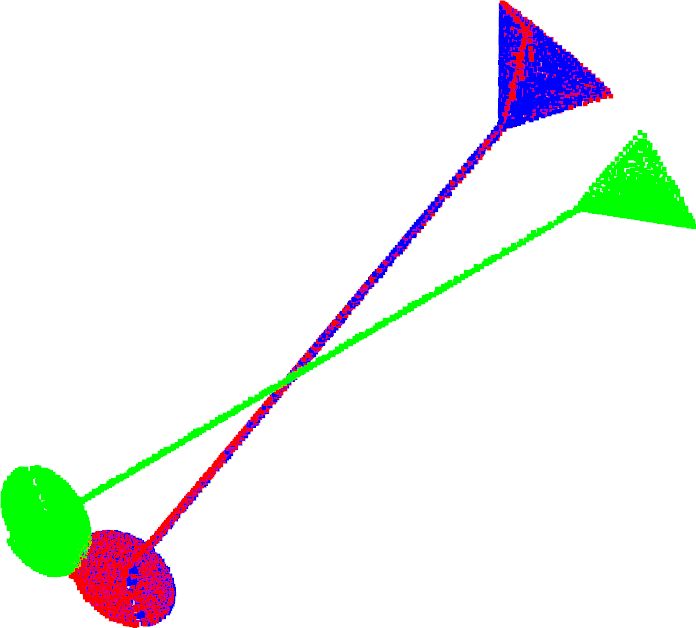} & \includegraphics[width=0.24\textwidth]{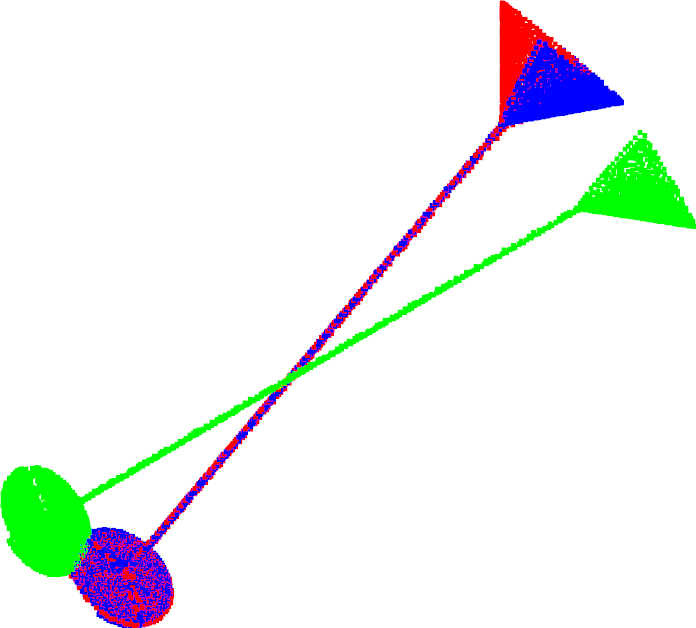}  \\  
(a1) 1-89-predator-180 & (a2) 1-89-ropnet-0.3 & (b1) 4-308-predator-176 & (b2) 4-308-ropnet-32 \\
\includegraphics[width=0.24\textwidth]{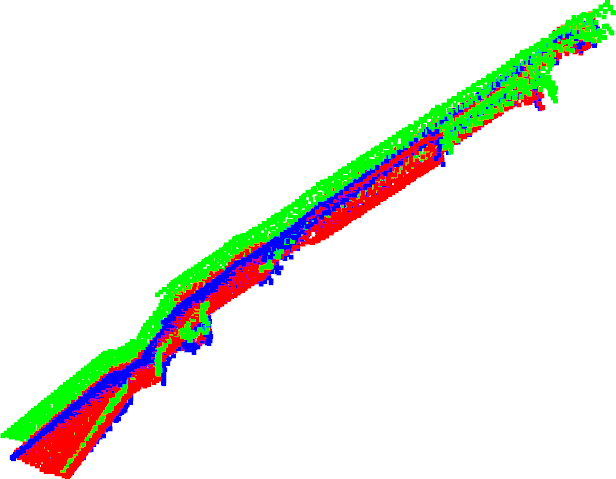} & \includegraphics[width=0.24\textwidth]{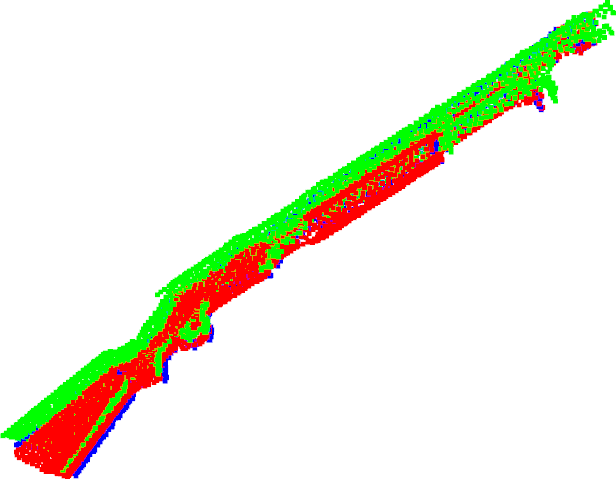} & \includegraphics[width=0.24\textwidth]{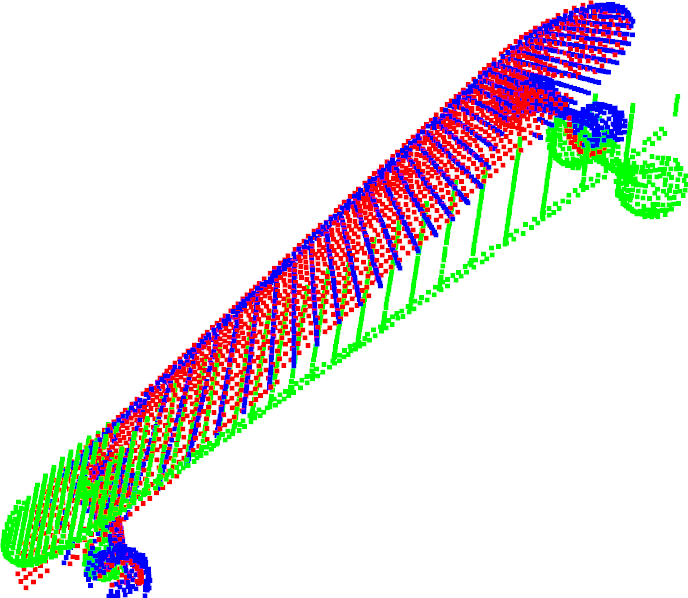} & \includegraphics[width=0.24\textwidth]{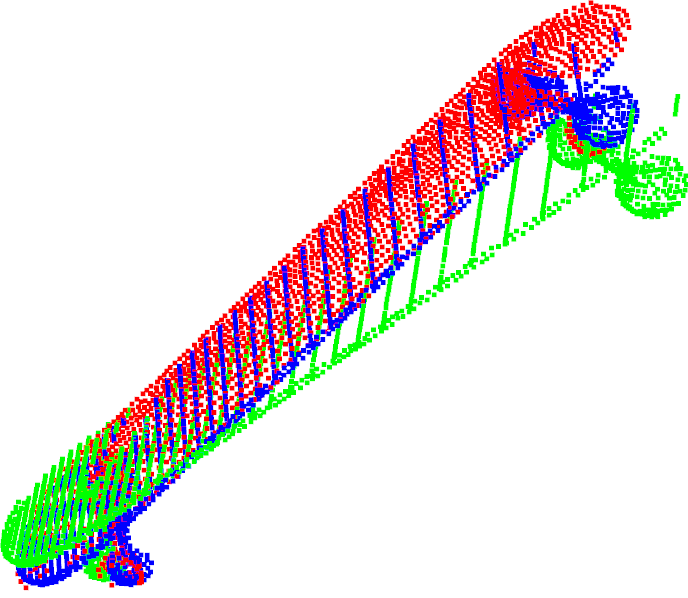}  \\  
(c1) 14-1081-predator-14 & (c2) 14-1081-ropnet-0.8 & (d1) 15-1155-predator-180 & (d2) 15-1155-ropnet-1 \\
\end{tabular}}
\caption{Visualization of registration on validation set.  The source, target and predicted point cloud are in green, red and blue, respectively. The description below denotes category-id-model-Error(R). 
}
\label{fig:data_vis}
\end{figure*}

We applied ensemble models based on the observation that some point cloud pairs in the MVP dataset are ambiguous and challenging, as shown in \autoref{fig:data_vis}:
\begin{itemize}
  \item (a1)(a2) denotes registration of plane-oriented categories.
  \item (b1)(b2) denotes registration of rotational-symmetry categories.
  \item (c1)(c2) denotes registration with very low overlap.
  \item (d1)(d2) denotes registration of axisymmetric categories.
  \item (a1)(b1)(d1) are categories which are ambiguous for registration.
\end{itemize}

For example, as shown in \autoref{fig:data_vis} (a1)(a2), we intuitively believe that the two registration results are reasonable.
However, we calculated the rotation error based on ground truth transformation, and obtain error of 180 with PREDATOR and error of 0.3 with ROPNet, respectively.
From \autoref{fig:data_vis} (b1)(b2) and (d1)(d2), we get the same conclusion. We think they all are ambiguous pairs.
For pair  \autoref{fig:data_vis} (c1) with low overlap, RANSAC implementation by Open3D~\cite{zhou2018open3d} v0.9 in PREDATOR, whose evaluation criteria is based on overlap, tends to obtain a higher overlap which it is not reasonable for pairs like (c1).
Besides, as shown in \autoref{fig:pipeline}, when sampled points are concentrated on the same area during each RANSAC iteration, they can hardly get the proper registration result.

Based on the above observations, we try to fuse ROPNet and PREDATOR on the MVP registration challenge.
Here, we designed four rules to select the final transformation based on overlap and rotation matrix.

First, let's explain some mathematical symbols. We reuse $T_1$ and $T_3$ defined in \autoref{sec:method}. 
Also, we estimate the transformation $T_2$ from target point cloud to source point cloud based on ROPNet. We also calculated overlap $OL_1$ and $OL_3$ between the transformed source point cloud and target point cloud based on $T_1$ and $T_3$. For each transformation $T_i, i \in \{1, 2, 3\}$, we have rotation matrix $R_i \in \mathbf{SO(3)}$ and translation vector $t_i \in \mathbb{R}^3$. Now, we have the following specially designed rules:
\begin{itemize}
\item{$l_1$: If rotation error between $R_1$ and $R_2^T$ is smaller than the defined threshold $d_1$, we select $T_1$ with high confidence.}
\item{$l_2$: If $R_1$ (in degree) is smaller than the threshold $d_2$, we select $T_1$ with high confidence.}
\item{$l_3$: If $OL_1$ is smaller than the threshold $d_3$, we select $T_3$ with high confidence}
\item{$l_4$: If $OL_1 + d_4 > OL_3$, we select $T_1$ with high confidence. $d_4$ is the pre-defined threshold}.
\end{itemize}

We set different $d_i, i \in \{1, 2, 3, 4\}$ for each category on the validation set. 
We fuse the above rules using the following predicate:  $$l = (l_1 \ \text{and} \  l_2 \  \text{and }\ l_3) \  \text{or} \ l_4$$ to decide which model we should use. 

If the result of evaluating $l$ is True, we select ROPNet, otherwise PREDATOR.
Experiments on the validation set showed it is effective on most pair cases, with few failed cases.

\item[$\bullet$] What was the benefit over the single method?

As observed in \autoref{fig:data_vis}, some pairs are ambiguous, with low overlap or under unrestricted rotation. ROPNet is good at processing the registration problems with small rotation angles, even when the pair has low overlap. However, it is difficult to deal with registration under large rotations. PREDATOR can solve registration problems under unrestricted angles. However, it is not ideal for object-centric point clouds registration, especially for pairs with low overlap,  plane structures or symmetric structures.
Considering the above observations, we fused the two models and achieved better performance in MVP registration track.

\item[$\bullet$] What were the baseline and the fused methods?

We can see the quantitative results in \autoref{table:single} and \autoref{table:ensemble-results}. More visualization results can be seen in \autoref{fig:results_vis}.

\begin{table}
\begin{center}
\scalebox{0.8}{
\begin{tabular}{|l|c|c|c|c|}
\hline
Model & rot\_level & Error(R) & Error(t) & MSE \\
\hline
ROPNet & 0 & 1.0449 & 0.0193 & 0.0375 \\
PREDATOR & 0 & 6.5910 & 0.0430 & 0.1581 \\
PREDATOR & 0, 1 & 6.8290 & 0.0413 & 0.1605 \\
\hline
\end{tabular}}
\end{center}
\caption{Evaluation on validation set based on single model.}
\label{table:single}
\end{table}

\begin{table*}
\begin{center}
\scalebox{0.7}{
\begin{tabular}{|c|c|c|c|c|c|c|c|c|c|c|c|c|c|c|c|c|c|c|}
\hline
label & 0 & 1 & 2 & 3 & 4 & 5 & 6 & 7 & 8 & 9 & 10 & 11 & 12 & 13 & 14 & 15 & total & total \\
\hline
type & \multicolumn{17}{|c|}{validation} & test \\
\hline

Error(R) & 0.4579 & 2.7508 & 0.6073 & 0.6676 & 8.2973 & 0.4111 & 5.3276 & 1.5040 & 5.1859 & 5.3121 & 8.7753 & 3.2399 & 0.2696 & 0.5272 & 1.5929 & 5.7384 & 3.16656 & 2.96546\\ 
Error(t) & 0.0198 & 0.0517 & 0.0152 & 0.0143 & 0.0392 & 0.0131 & 0.0855 & 0.0281 & 0.0258 & 0.0305 & 0.0434 & 0.0255 & 0.0115 & 0.0111 & 0.0192 & 0.0339 & 0.029237 & 0.02632\\
MSE & 0.0278 & 0.0997 & 0.0258 & 0.0260 & 0.1840 & 0.0202 & 0.1785 & 0.0544 & 0.1163 & 0.1233 & 0.1966 & 0.0820 & 0.0162 & 0.0203 & 0.0470 & 0.1341 & 0.08451 & 0.07808\\
\hline
\end{tabular}}
\end{center}
\caption{Evaluation on validation and test set based on ensemble model.}
\label{table:ensemble-results}
\end{table*}

\begin{figure*}
\scalebox{1.0}{
\begin{tabular}{cc|cc|cc|cc}
\includegraphics[width=0.1\textwidth]{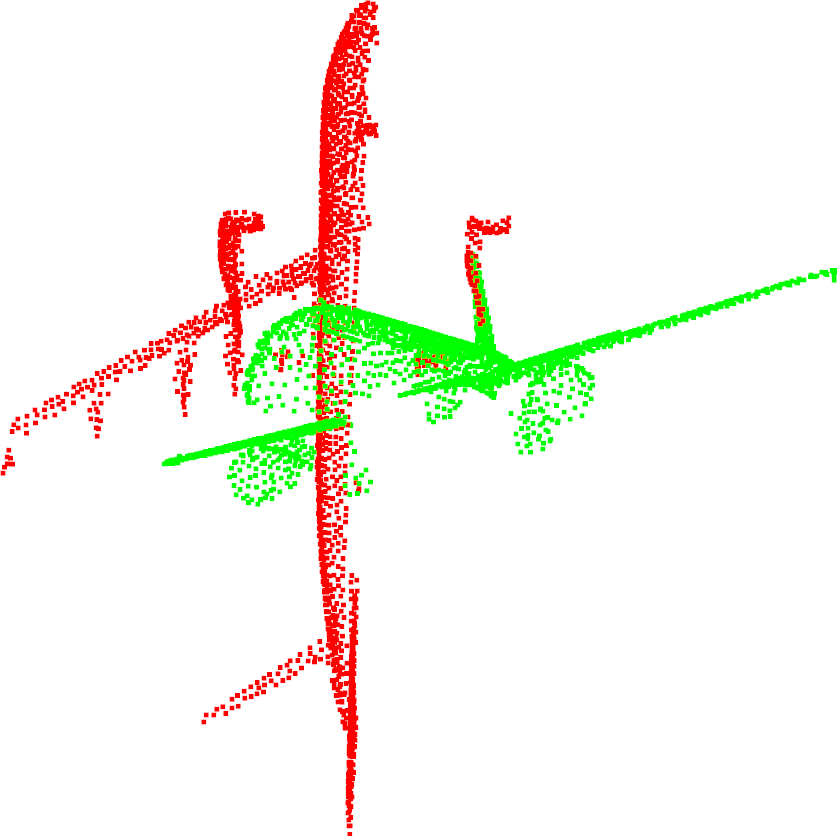} & \includegraphics[width=0.1\textwidth]{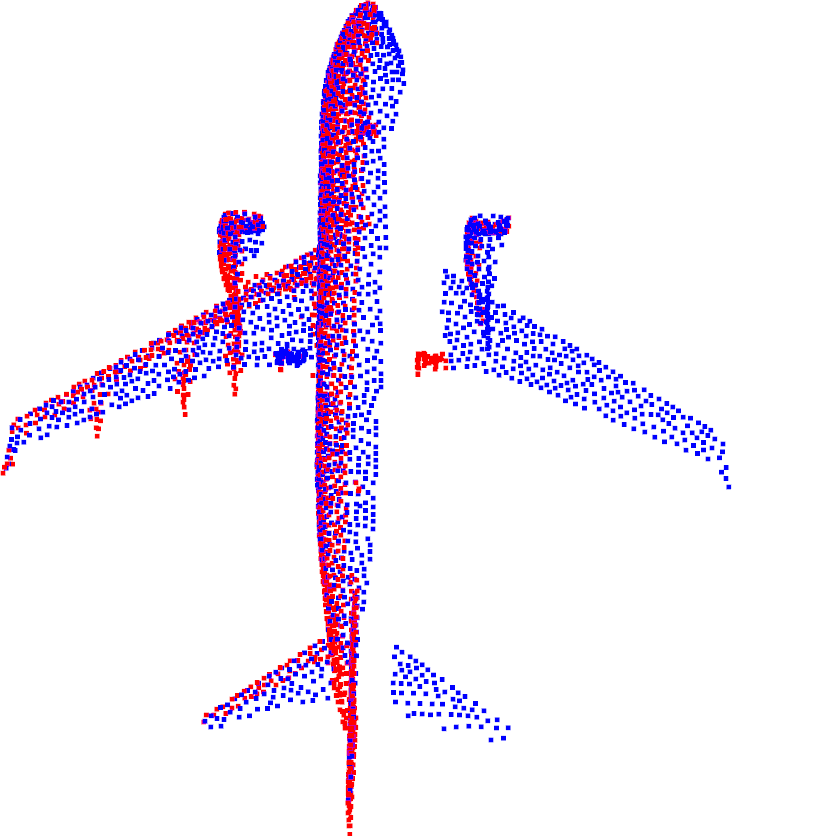} & \includegraphics[width=0.1\textwidth]{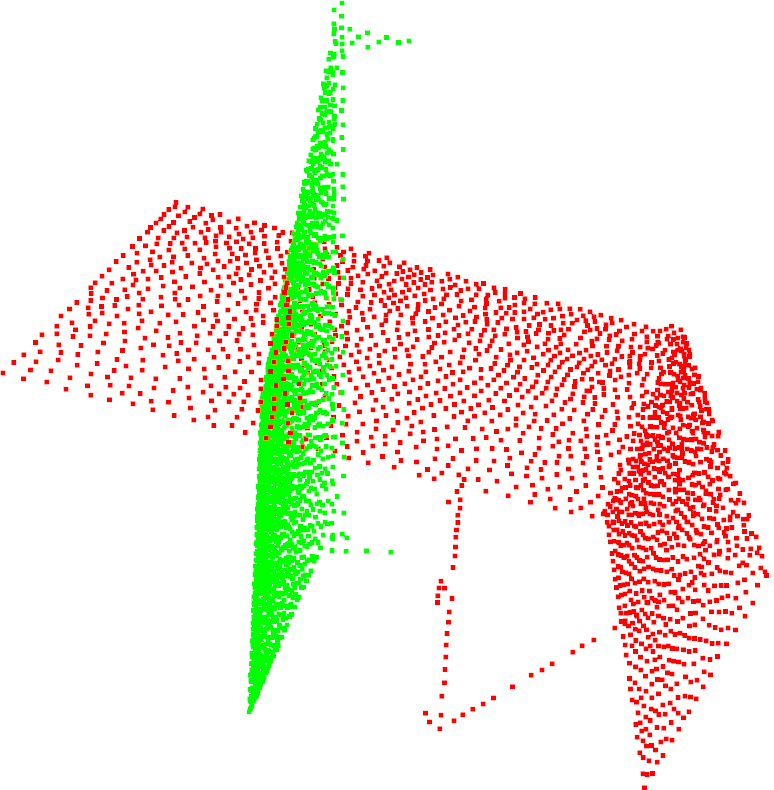} & \includegraphics[width=0.1\textwidth]{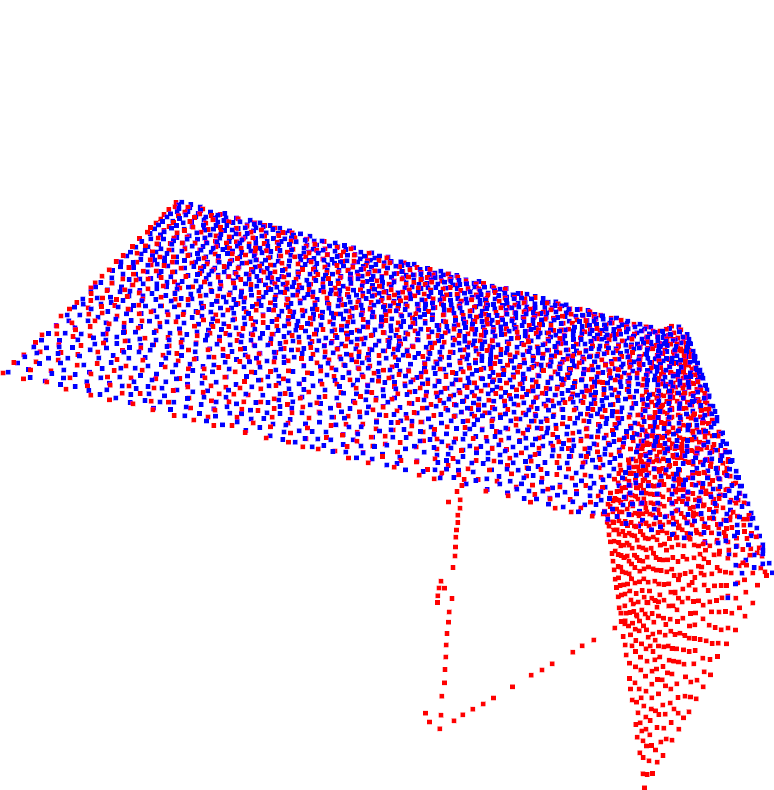} & \includegraphics[width=0.1\textwidth]{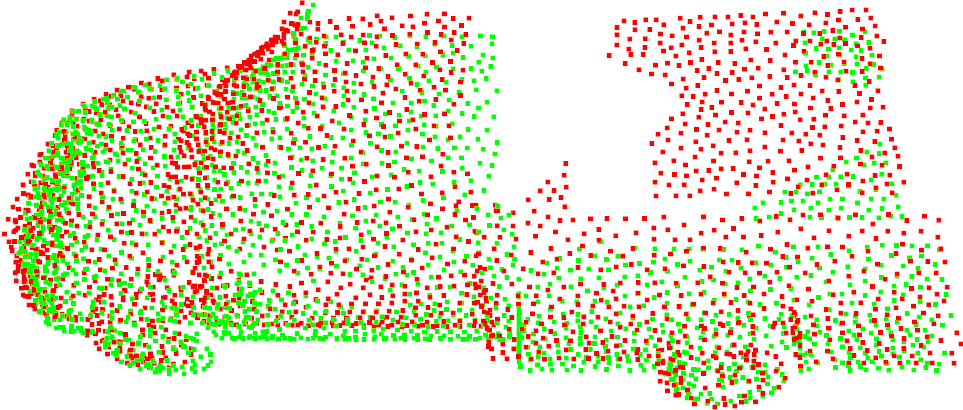} & \includegraphics[width=0.1\textwidth]{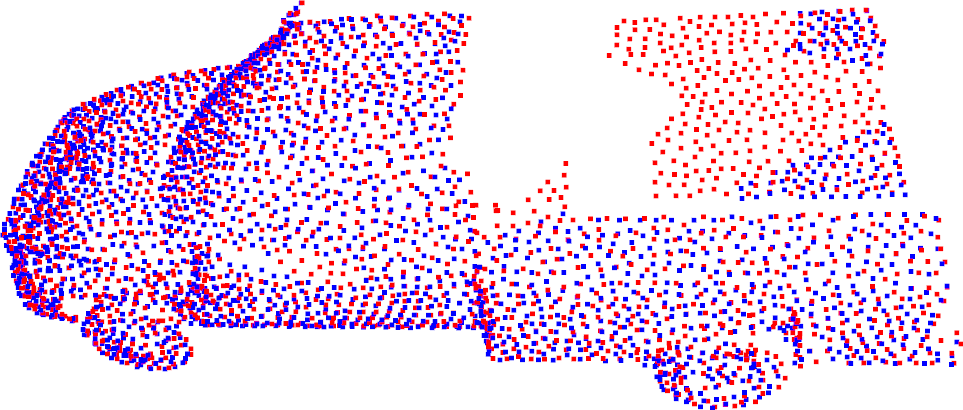} & \includegraphics[width=0.1\textwidth]{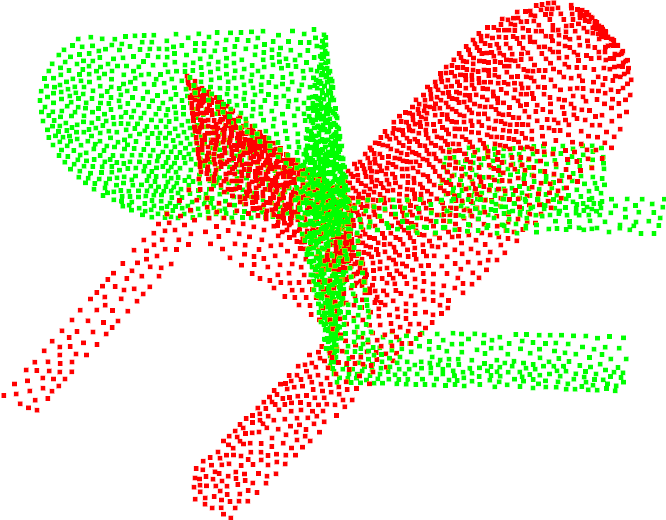} & \includegraphics[width=0.1\textwidth]{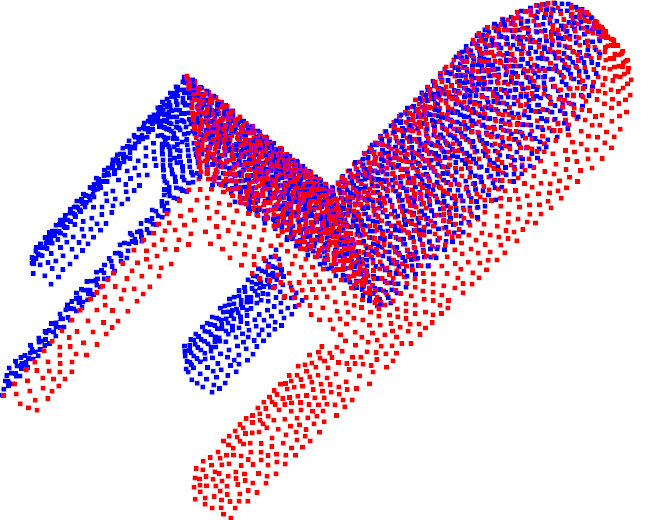}  \\  
\includegraphics[width=0.1\textwidth]{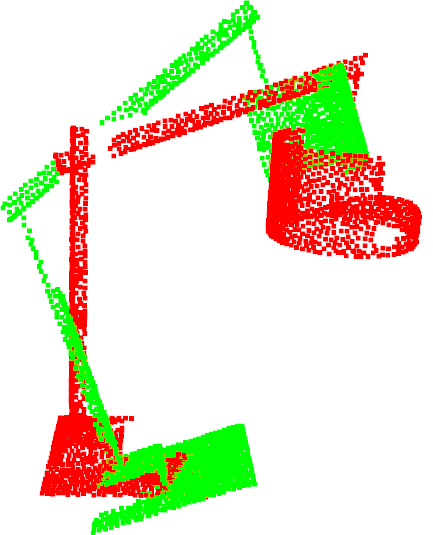} & \includegraphics[width=0.1\textwidth]{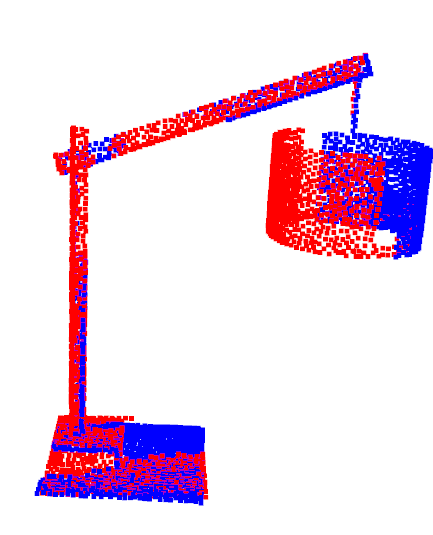} & \includegraphics[width=0.1\textwidth]{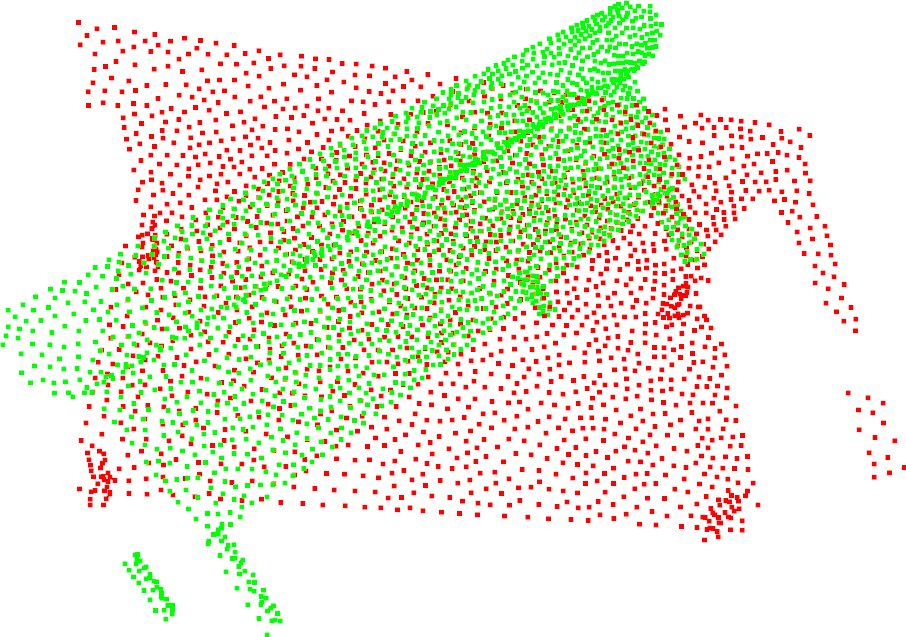} & \includegraphics[width=0.1\textwidth]{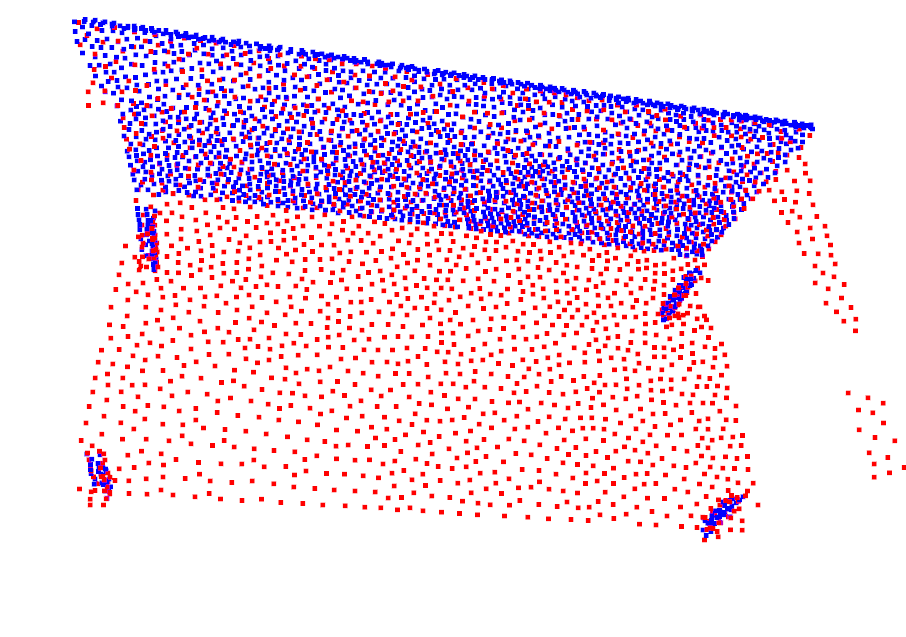} & \includegraphics[width=0.1\textwidth]{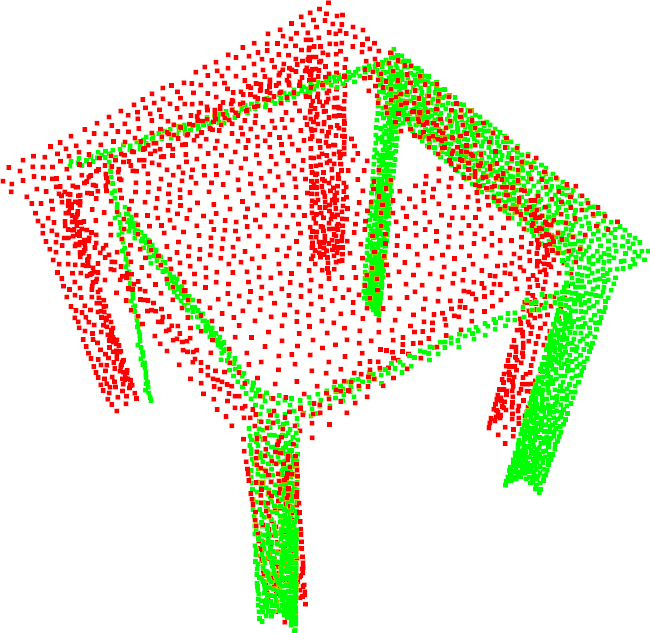} & \includegraphics[width=0.1\textwidth]{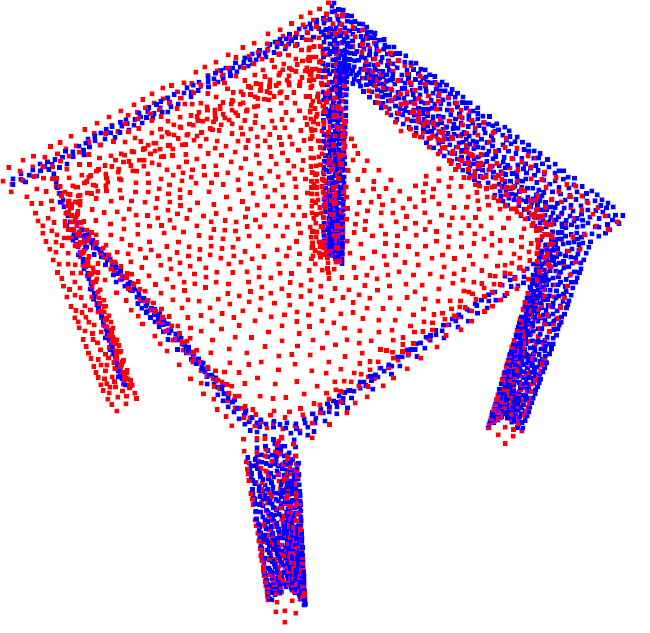} & \includegraphics[width=0.1\textwidth]{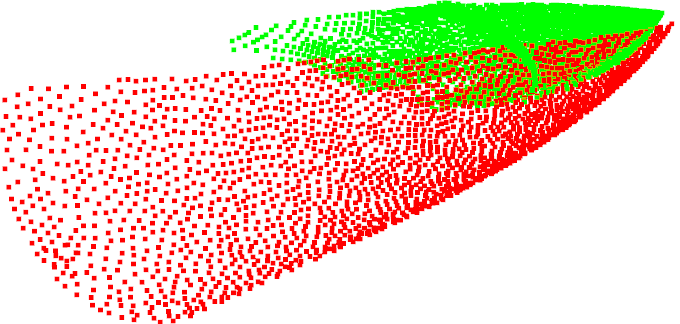} & \includegraphics[width=0.1\textwidth]{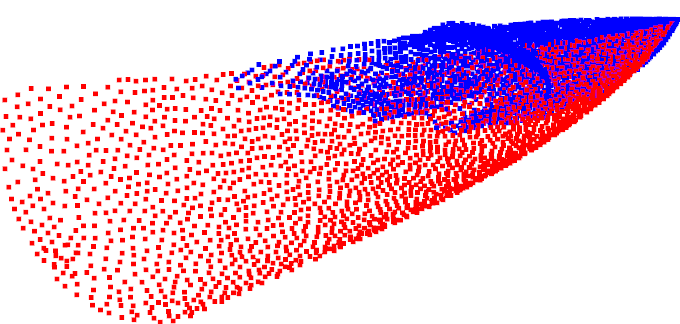}  \\ 
\includegraphics[width=0.1\textwidth]{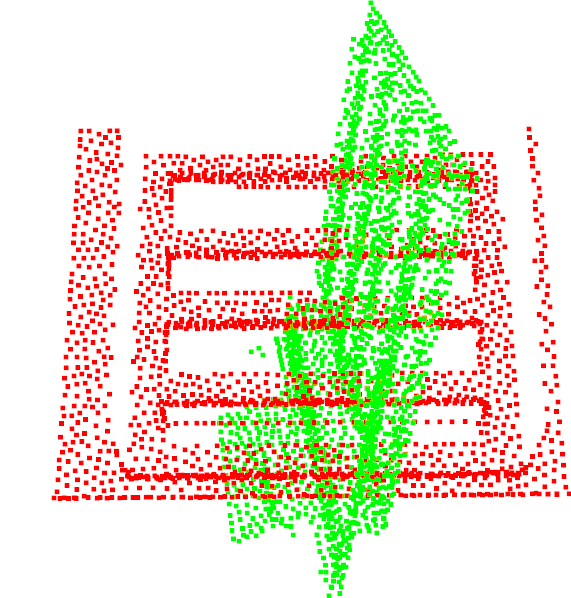} & \includegraphics[width=0.1\textwidth]{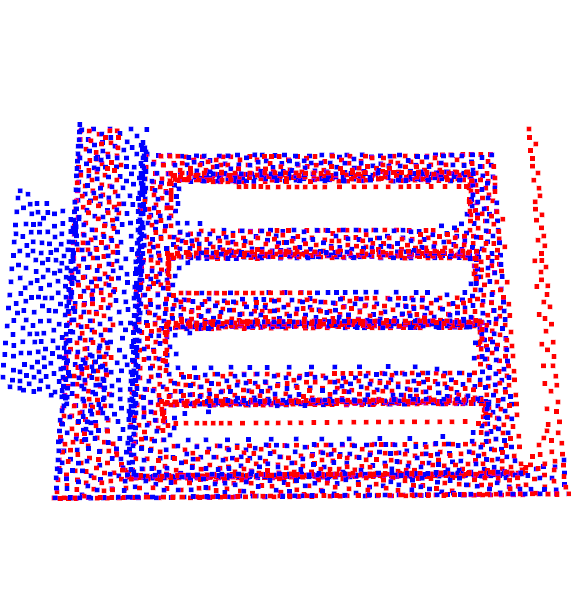} & \includegraphics[width=0.1\textwidth]{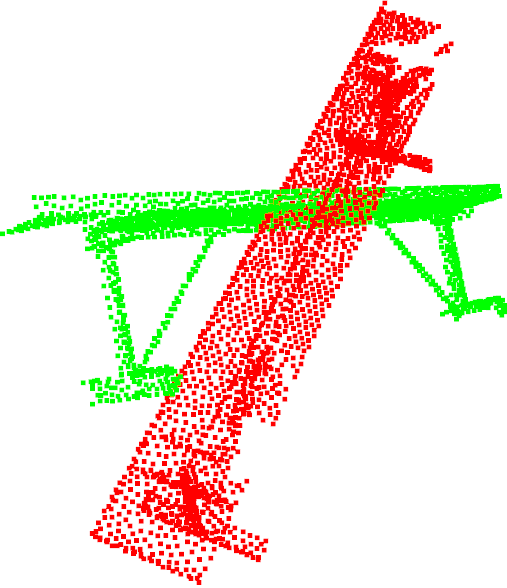} & \includegraphics[width=0.1\textwidth]{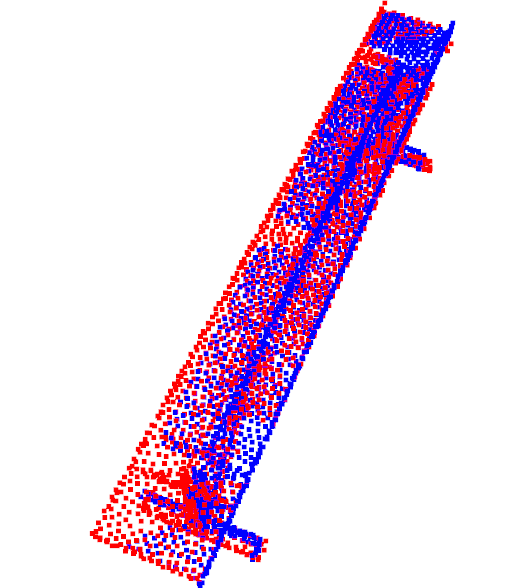} & \includegraphics[width=0.1\textwidth]{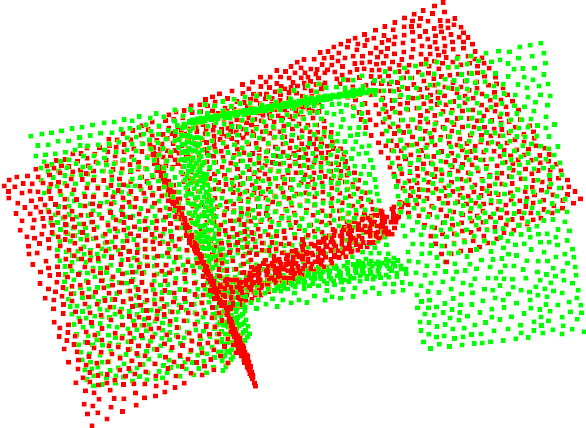} & \includegraphics[width=0.1\textwidth]{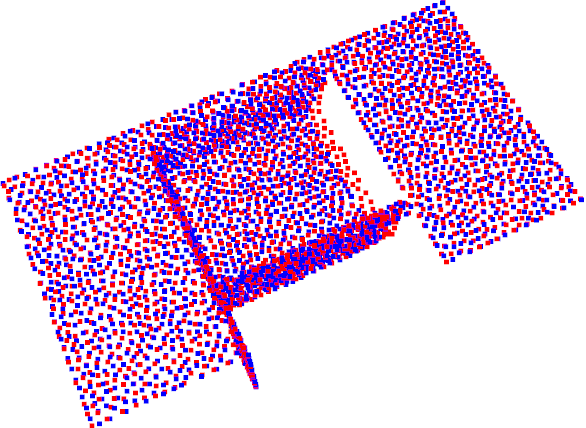} & \includegraphics[width=0.1\textwidth]{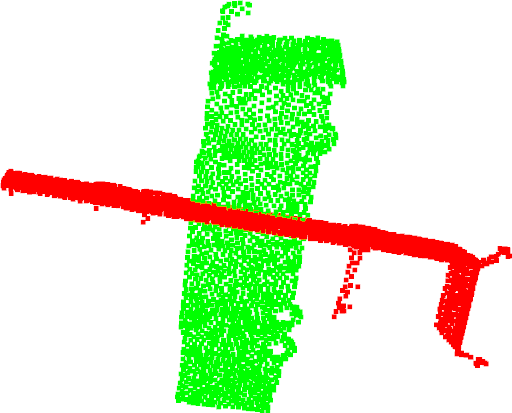} & \includegraphics[width=0.1\textwidth]{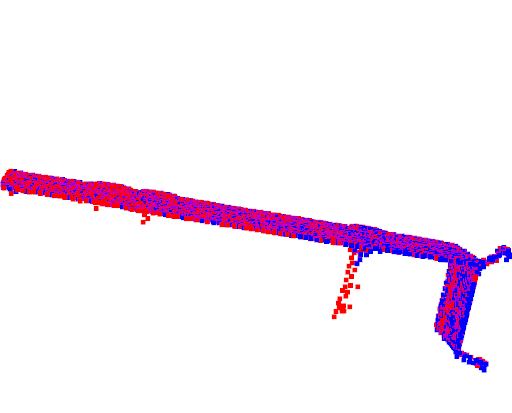}  \\ 
\includegraphics[width=0.1\textwidth]{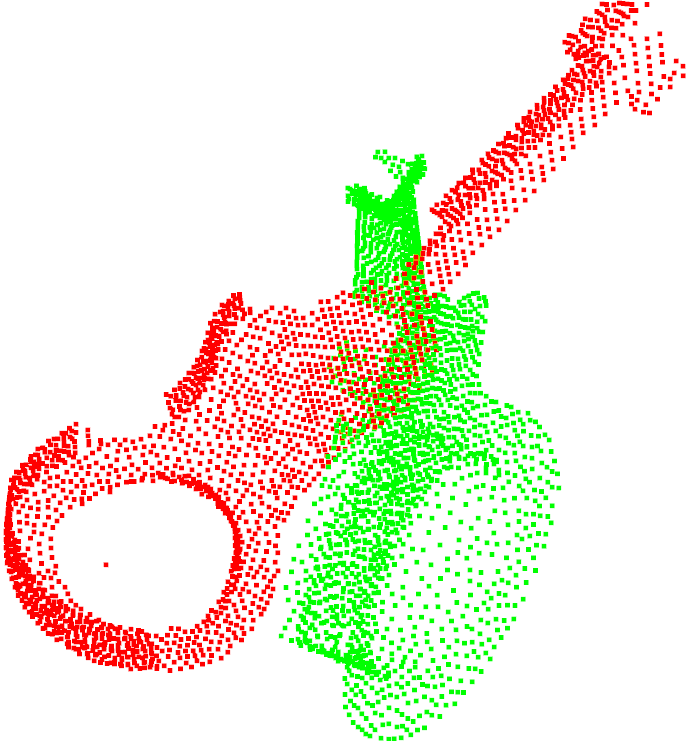} & \includegraphics[width=0.1\textwidth]{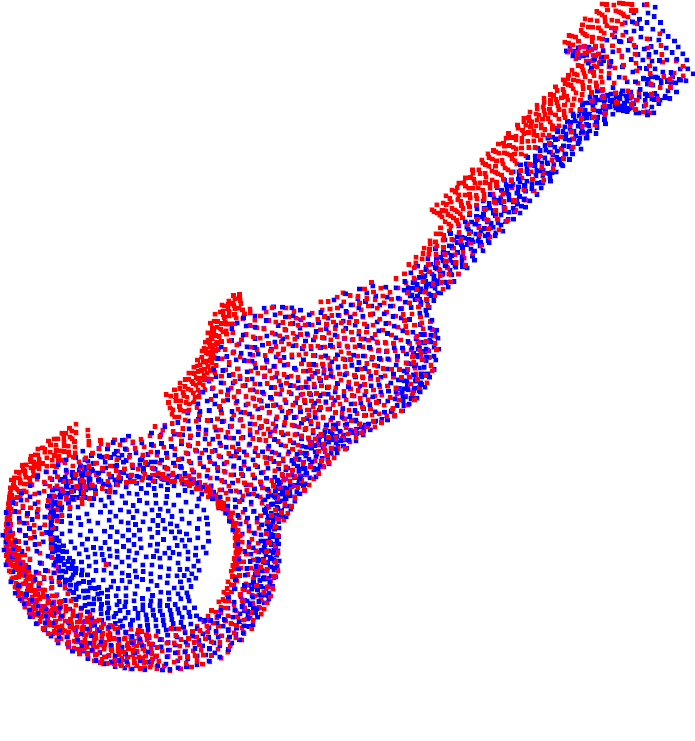} & \includegraphics[width=0.1\textwidth]{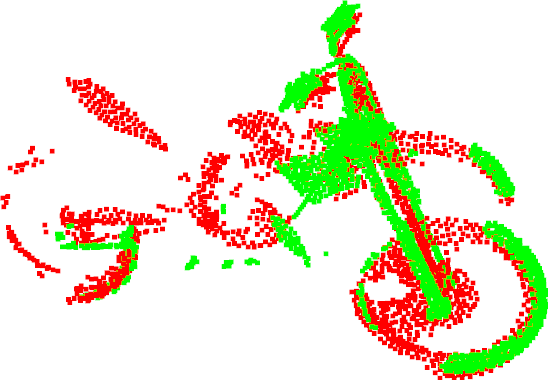} & \includegraphics[width=0.1\textwidth]{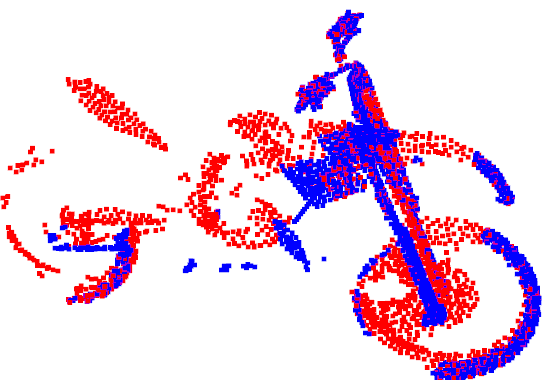} & \includegraphics[width=0.1\textwidth]{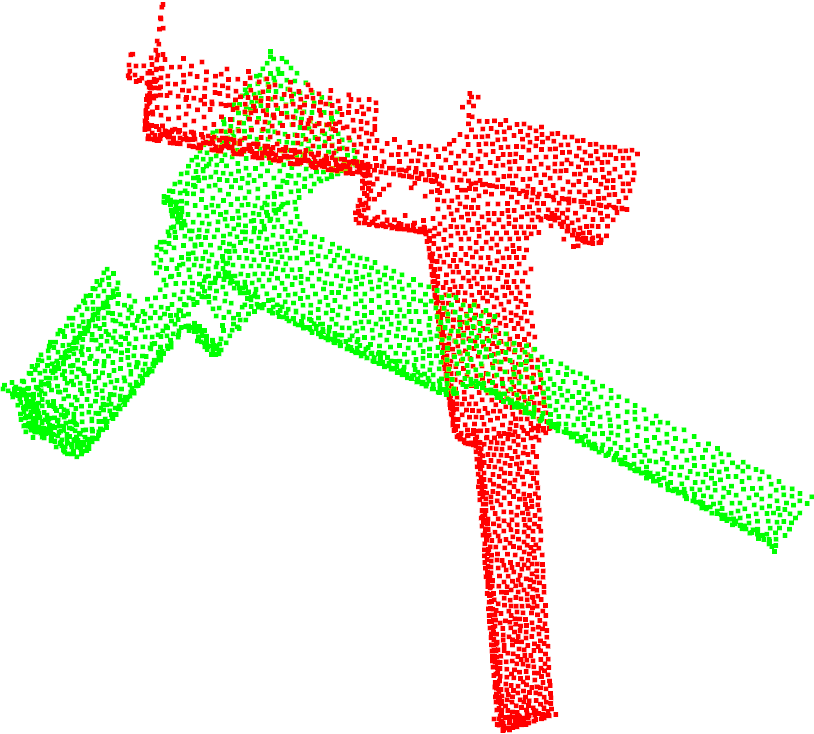} & \includegraphics[width=0.1\textwidth]{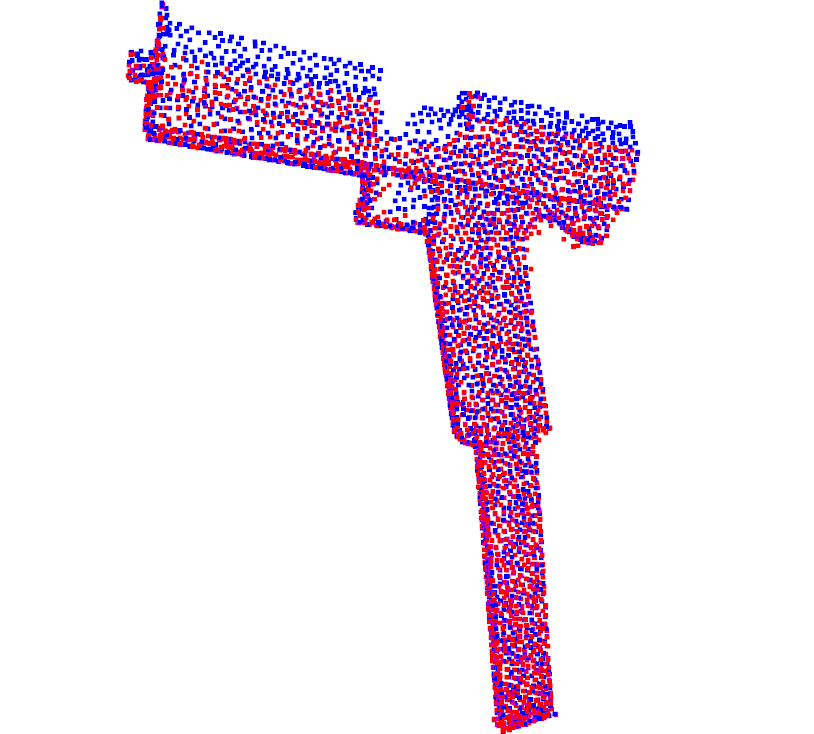} & \includegraphics[width=0.1\textwidth]{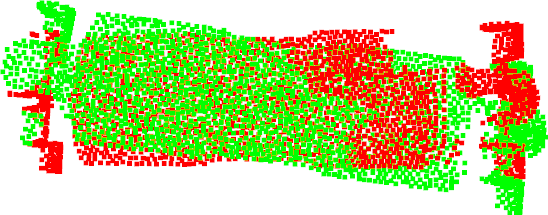} & \includegraphics[width=0.1\textwidth]{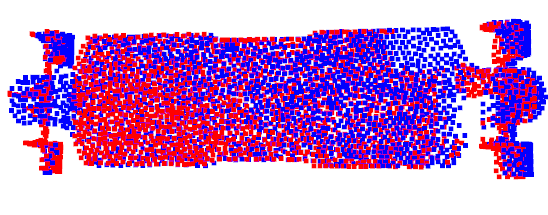} \\
\includegraphics[width=0.1\textwidth]{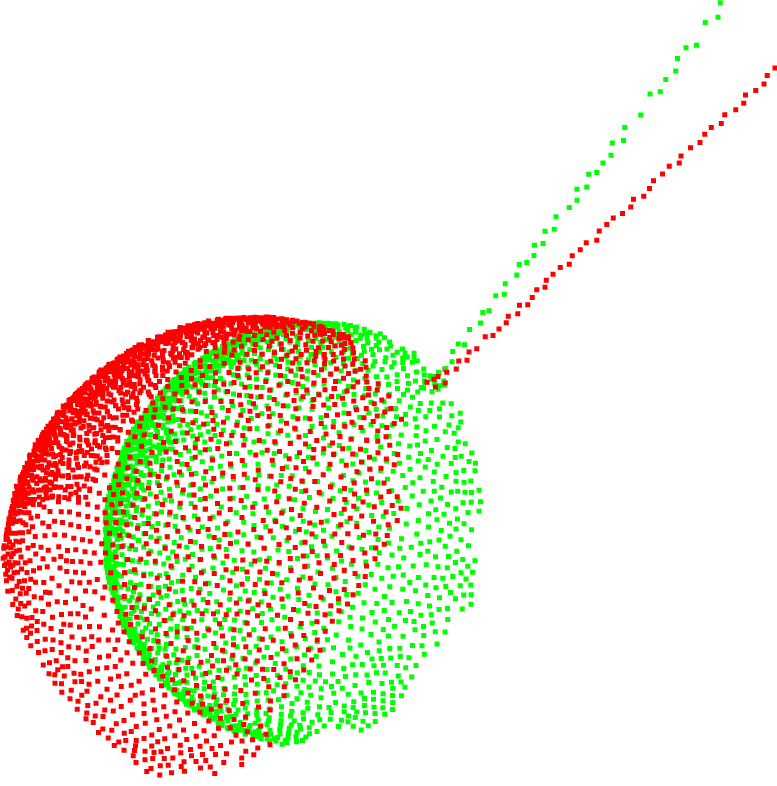} & \includegraphics[width=0.1\textwidth]{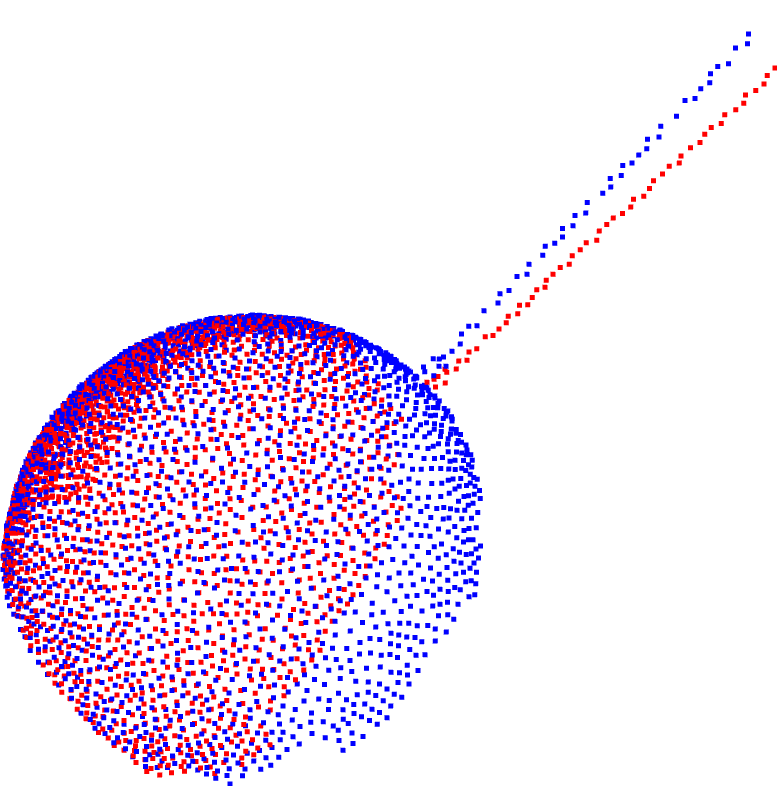} &
\includegraphics[width=0.1\textwidth]{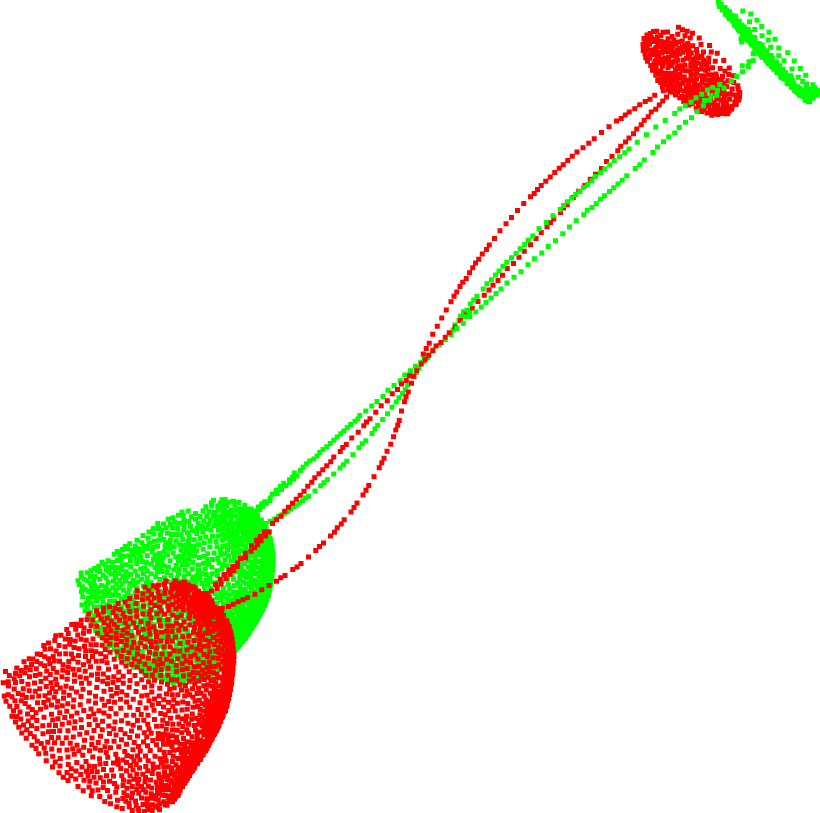} & \includegraphics[width=0.1\textwidth]{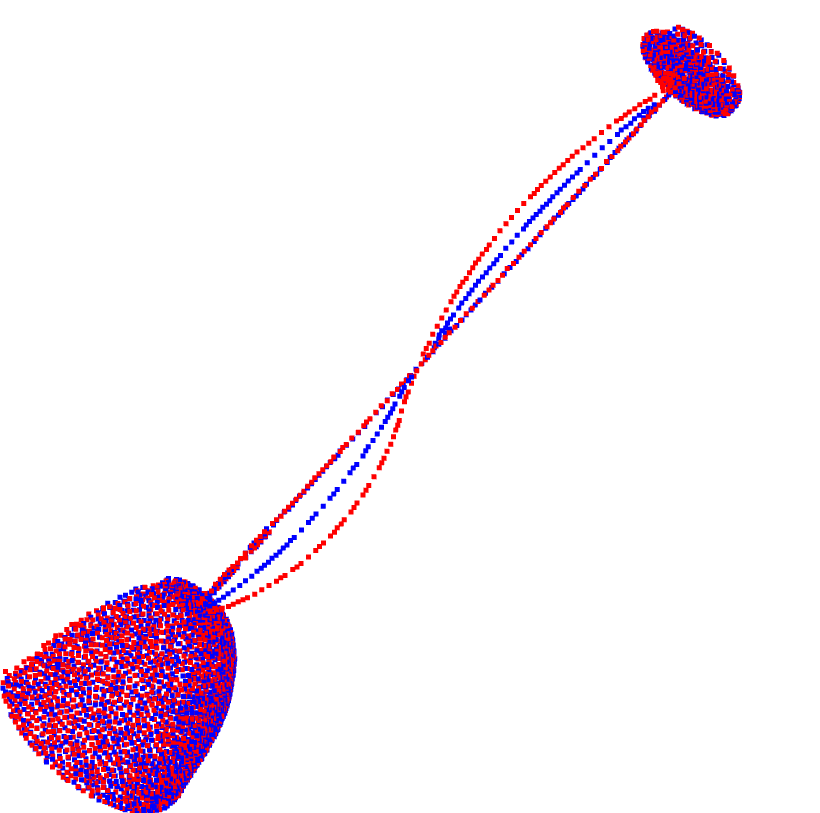} & \includegraphics[width=0.1\textwidth]{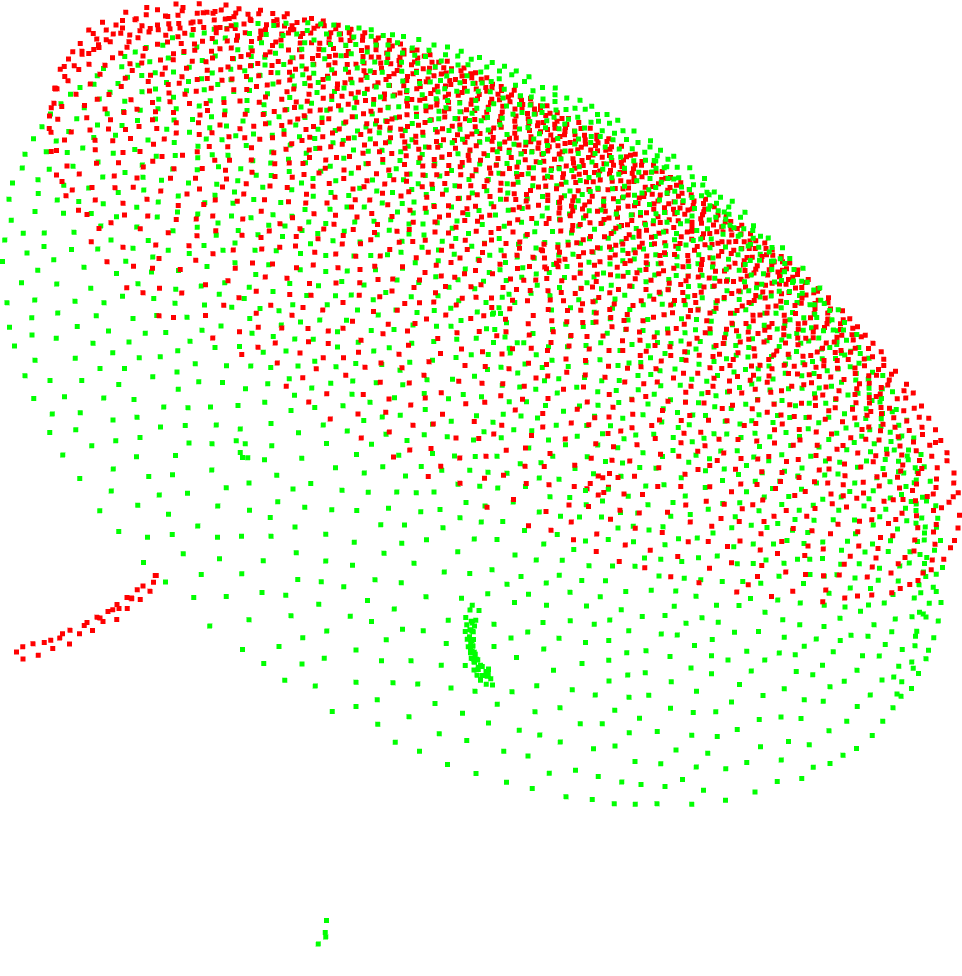} &  \includegraphics[width=0.1\textwidth]{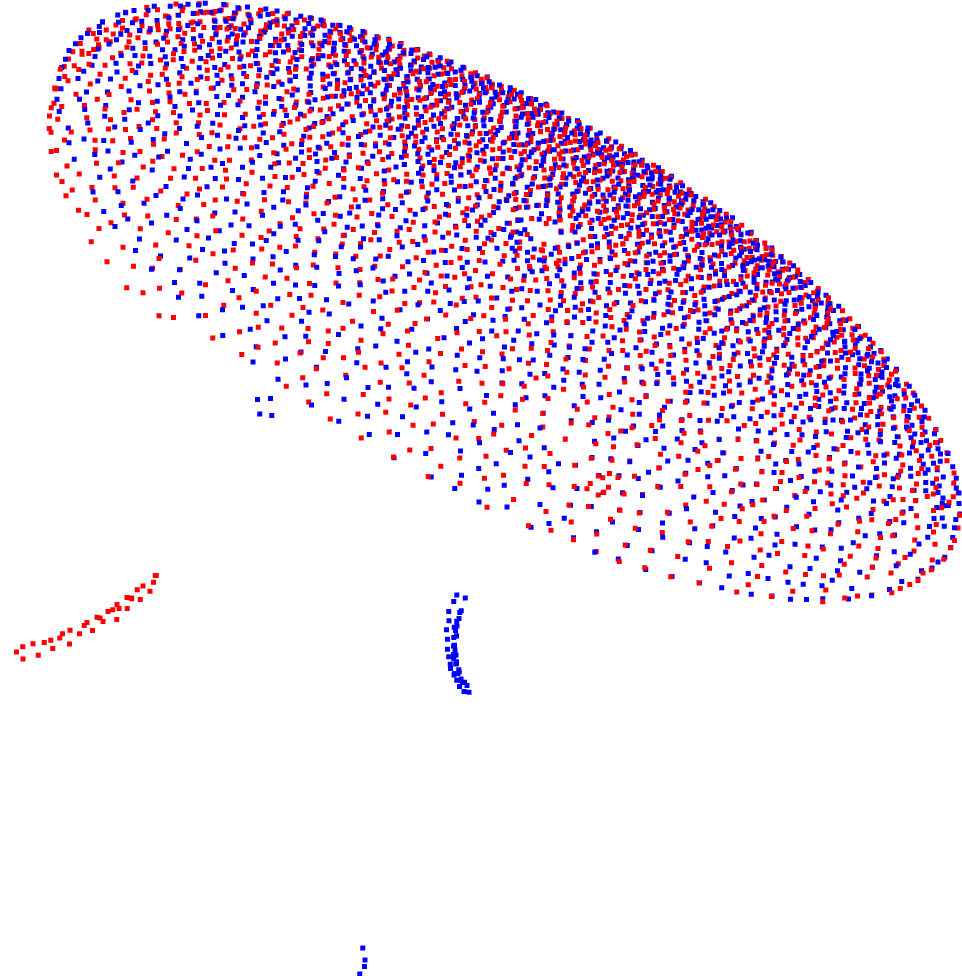} & \includegraphics[width=0.1\textwidth]{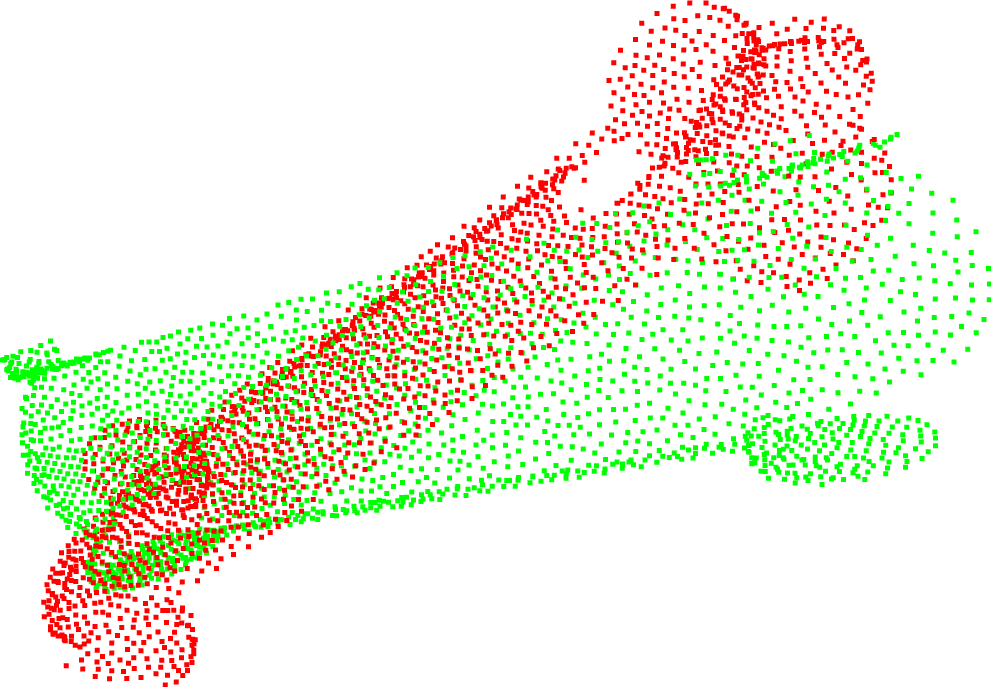} & \includegraphics[width=0.1\textwidth]{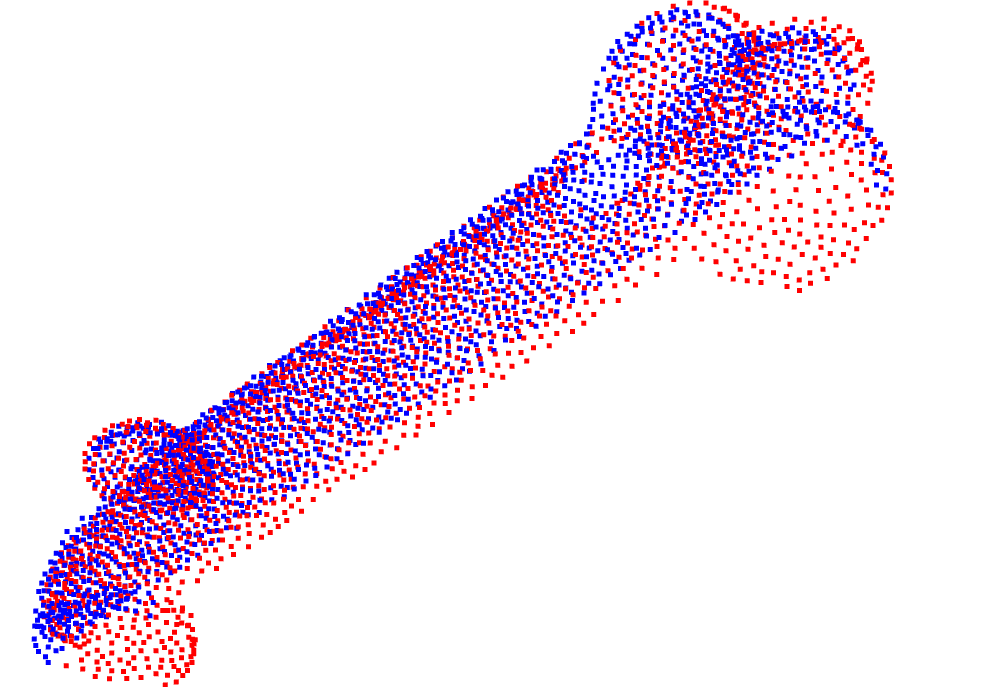}

\end{tabular}
}
\caption{Registration visualization on MVP registration. The first 4 rows show registration results on test set. The last row shows the registration results on validation set whose rotation errors (in degree) are bigger than 10°. The source and target point cloud are marked in green and red respectively. The blue one is the transformed point cloud using the estimating transformation.}
\label{fig:results_vis}
\end{figure*}

\end{itemize}

\section{Reproducibility Details}
\begin{itemize}
\item[$\bullet$] Implementation details (including language, platform, parallelization and memory requirements)

We have tested our code on Ubuntu 16.4, Python 3.7, PyTorch (1.7.1+cu101) torchvision (0.8.2+cu101), GCC 5.4.0, Open3D 0.9.0. Almost all experiments were run on a Tesla V100 GPU with an Intel 6133 CPU @ 2.50GHz, 320G RAM. Part of test experiments were run on a single GeForce GTX TITAN X with Intel(R) Core(TM) i7-6700K CPU @ 4.00GHz, 32GB RAM.

 

\item[$\bullet$] Training/testing time?

It required about 7 minutes to train ROPNet for one epoch on MVP registration data (with 2048 points), and about 70 hours for training 600 epochs. It required about 18 minutes to train PREDATOR for one epoch, and about 60 hours for training 200 epochs. The training speed was tested on shared server while many other tasks were also using the GPU, CPU and memory, so training may be faster for a clean server machine.
Finally, it required 0.768s to obtain the registration result for a pair of point clouds with 2048 points. More details about testing time can be seen in ~\autoref{table:speed}.

\end{itemize}

\section{Feedback}
\begin{itemize}

\item[$\bullet$] General comments on the MVP Challenge 2021.

We found participating in the Point cloud registration track on the MVP Challenge 2021 was very interesting and challenging. The data set considers different overlap, density and unrestricted rotation angles, having several differences with ModelNet40 which is used as registration benchmark in many recent works.
We learned a lot from the competition and believe it will help us to design more robust registration algorithms based on multiple consideration.

Unfortunately, I am a little disappointed with one thing: the data quality and metrics. 
As we observed in the validation set, there are some pairs with just plane objects (category 1), or symmetrical objects (categories 4, 6, 9, 15). Some ground truth results  are not reasonable under the metrics used in this challenge, as we described in \autoref{sec:ensemble}. 
However, by re-cleaning the data set and improving the evaluation metrics, we think this issue can be solved, and a new and unified point cloud registration benchmark can be produced for the community.

\item[$\bullet$] What do you expect on a new competition on point cloud related tasks?

In our opinion, and mentioned before, re-cleaning the data set, improving the evaluation metrics, and then setting up a new, unified point cloud registration benchmark (like 3DMatch~\cite{zeng20173dmatch}, KITTI~\cite{geiger2012we}) for registration community is a good choice. As far as we know, some point cloud registration works are not compared fairly on ModelNet40, because they may use different sampled points, different transformation matrices and different partial point clouds generation. Besides, the overlap size, rotation angles and points density are set in easier configurations.

\item[$\bullet$] Other comments (if any): encountered difficulties, proposed tracks, proposed evaluation metric(s), proposed challenge platform, etc.

Evaluation metrics should be improved, such as considering  chamfer distance between the transformed point cloud and complete point cloud, or evaluating symmetric and asymmetric objects registration performance respectively.

Also, there is a small shortcoming in the platform, in which each user can hide their submissions or best result. 

\end{itemize}

{\small
\bibliographystyle{ieee_fullname}
\bibliography{egbib}
}

\end{document}